\documentclass[journal]{IEEEtran}
\usepackage{amssymb}
\usepackage{times}
\usepackage{epsfig}
\usepackage{graphicx}
\usepackage{amsmath}
\usepackage{algorithm}
\usepackage{algorithmic}
\usepackage{subfig}
\usepackage{multirow}
\usepackage{color}
\usepackage{caption}
\usepackage{cite}
\usepackage{mathptmx}  
\usepackage{mathrsfs}
\usepackage{amsthm}
\usepackage{amsfonts}
\usepackage{url} 
\usepackage{textcomp}
\usepackage[table]{xcolor}
\usepackage[pagebackref=true,breaklinks=true,letterpaper=true,colorlinks,bookmarks=false,citecolor=blue]{hyperref}

\ifCLASSINFOpdf
\else
\fi
\begin{document}
%
\title{PTB-TIR: A Thermal Infrared Pedestrian Tracking Benchmark }
%
%
%

\author{Qiao~Liu,
        Zhenyu~He,~\IEEEmembership{Senior Member,~IEEE,} Xin~Li, Yuan~Zheng

\thanks{Q. Liu, Z. He (Corresponding author) and X. Li are with the School of Computer Science and Technology, Harbin Institute of Technology, Shenzhen, China. e-mail: (zhenyuhe@hit.edu.cn).}
\thanks{Yuan Zheng is with the School of Computer Science, Inner Mongolia University, Huhehot, China.}
\thanks{Supplementary material provides more details of the dataset and experimental results, which is available at~\url{https://ieeexplore.ieee.org/document/8798885/media\#media}.}
\thanks{DOI: 10.1109/TMM.2019.2932615.}
}

%
%

\markboth{}%
{Shell \MakeLowercase{\textit{et al.}}: Bare Demo of IEEEtran.cls for IEEE Journals}
%



\maketitle

\begin{abstract}
Thermal infrared (TIR) pedestrian tracking is one of the important components among numerous applications of computer vision, which has a major advantage: it can track pedestrians in total darkness.  The ability to evaluate the TIR pedestrian tracker fairly, on a benchmark dataset, is significant for the development of this field. However, there is not a benchmark dataset. In this paper, we develop a TIR pedestrian tracking dataset for the TIR pedestrian tracker evaluation. The dataset includes 60 thermal sequences with manual annotations. Each sequence has nine attribute labels for the attribute based evaluation. In addition to the dataset, we {carry} out the large-scale evaluation experiments on our benchmark dataset using nine publicly available trackers. The experimental results help us understand the strengths and weaknesses of these trackers.
In addition, in order to gain more insight into the TIR pedestrian tracker, we {divide} its functions into three components: feature extractor, motion model, and observation model. Then, we conduct three comparison experiments on our benchmark dataset to validate how each component affects the tracker's performance. The findings of these experiments provide some guidelines for future research.
\end{abstract}

\begin{IEEEkeywords}
thermal infrared, pedestrian tracking, benchmark, dataset
\end{IEEEkeywords}

%
\IEEEpeerreviewmaketitle

\section{Introduction}
\label{introduction}
\IEEEPARstart{A}s thermal imaging technology has developed, both the quality and the resolution of thermal images have been improved to a great extent. That enables a series of computer vision tasks based on thermal images to be applied in various fields.
Thermal infrared (TIR) pedestrian tracking is one important vision technology and has received significant attention in recent years. It {has a wide range of} applications such as traffic surveillance, driving assistance, and night rescue~\cite{TC,kwak2017pedestrian,ge2009real}.
Numerous TIR pedestrian trackers~\cite{xu2005pedestrian,li2010real,wang2012pedestrian,ko2013human,li2014robust,portmann2014people,ma2016pedestrian,yang2017pedestrian} have been proposed over the past decade. Despite substantial advancements, TIR pedestrian tracking still faces many challenges, e.g., occlusion, background clutter, motion blur, low resolution, and thermal crossover.

Evaluating and comparing different TIR pedestrian trackers fairly is crucial for the development of this field. Usually, there is not a single tracker that can handle all challenges. Therefore, it is necessary to compare and analyze the strengths and weaknesses of each tracker, thereby giving some guidelines to future research aimed at developing a better tracker. In order to do that, it is critical to collect a representative dataset.  Several TIR datasets have been used, such as the OSU Color-Thermal~\cite{OSU}, Terravic Motion IR~\cite{OTCBVS}, PDT-ATV~\cite{PDT-ATV}, and BU-TIV~\cite{TIV} databases. However, the TIR pedestrian trackers cannot be fairly compared on these datasets for the following reasons. First, these datasets only have a few sequences, and thus are not sufficiently large for a fair evaluation. Second, the thermal images are captured from a {singe thermal camera} that lacks diversity. Third, the thermal images have low resolution and static backgrounds that not suited to various real-world scenarios. Fourth, these datasets do not provide a unified ground-truth and attribute annotations. All of these reasons show that this field lacks a benchmark dataset to fairly evaluate the TIR pedestrian tracker.

The modern TIR pedestrian tracker is a complicated system, which consists of several separate components. Each component can affect its performance.
Evaluating a TIR pedestrian tracker as a whole helps us understand its overall performance, but the effectiveness of each component cannot be known.
To understand the TIR pedestrian tracker more sufficiently, it is necessary to evaluate each component separately.
Similarly to~\cite{wang2015understanding}, we divide the TIR pedestrian tracker into three components: feature extractor, motion model, and observation model.
After dividing it into these components, an interesting question will naturally arise. How do feature extractors, motion models, and observation models affect tracking performance respectively?
The answers to the question provide some useful information to guide future research.

\renewcommand\arraystretch{1.5}
\begin{table*}[htbp]
\centering
\caption{Comparison of our dataset with other datasets.}
\rowcolors{2}{gray!25}{white}
\begin{tabular}{c|ccccc}
  \hline
  \rowcolor{gray!50}
   & OSU Color-Thermal~\cite{OSU} & Terravic Motion IR~\cite{OTCBVS} & PDT-ATV~\cite{PDT-ATV} & BU-TIV~\cite{TIV} & PTB-TIR (\textbf{Ours}) \\
  \hline
   Device Type        & Raytheon PalmIR 250D & Raytheon Thermal-Eye 2000AS & FLIR Tau 320 & FLIR SC8000   & More than 8 sources \\
   Resolution      &$320\times 240$ & $320\times 240$ & $324\times 256$ & up to $1024\times 640$  & up to $1280\times 720$ \\
   Bit Depth        & 8       & 8      & 8     & 16   & 8 \\
   Sequences Number & 6   & 11  & 8 & 5 & 60 \\
  {Minimum Length} &   {601}     &  {360}    & {77} & {1000} &   {50}\\
  { Maximum Length} &   {2031}    & {1620}     & {775}   & {8920} &   {1451} \\
   Total Frames    & 8544 & 5500 & 3888 & 22654  & 30128 \\
  \hline
\end{tabular}\label{comd}
\end{table*}

In this paper, to fairly evaluate and compare TIR pedestrian trackers, we first collect a TIR pedestrian
dataset\footnote{The PTB-TIR dataset and code library can be accessed at~\url{http://www.hezhenyu.cn/PTB-TIR.html} or ~\url{https://github.com/QiaoLiuHit/PTB-TIR_Evaluation_toolkit}} for a short-term tracking task. The dataset includes $60$ video sequences with manual annotations. The entire dataset is divided into nine different attribute subsets. On each subset, we evaluate the tracker's ability in handling the corresponding challenge.  Then, we carry out a large-scale fair performance evaluation on nine publicly available trackers. Finally, in order to get insight into TIR pedestrian tracking, we further conduct three validation experiments on three components of the tracker. (1) We {compare} several different features on {two baseline trackers to analyze how different feature extractors affect tracking performance}. (2) We compare several different motion models on a baseline tracker to analyze how they affect tracking results. (3) We compare several trackers with different observation models to explore how they affect tracking performance. The findings of these three experiments allow us to understand the TIR pedestrian tracker more sufficiently.

The contributions of the paper are two-fold:
\begin{itemize}
  \item We {construct} a TIR pedestrian tracking benchmark dataset with $60$ annotated sequences for the TIR pedestrian tracker evaluation. A large-scale performance evaluation is implemented on our benchmark with nine publicly available trackers.
  \item Three validation experiments on the proposed benchmark are carried out to provide insight into the TIR pedestrian tracking system. The experimental results are analyzed to provide some guidelines for future research.
\end{itemize}

The rest of the paper is organized as follows. We first introduce TIR pedestrian tracking datasets and methods briefly in Section~\ref{RW}. Then, we present the contents of the proposed benchmark in Section~\ref{benchmark}. Subsequently, the extensive performance evaluation and validation experiments are reported in Section~\ref{exp1} and Section~\ref{exp2} respectively. Finally, we draw a short conclusion and describe some future work in Section~\ref{conclusion}.

\section{Related Work}
\label{RW}
In this section, we first introduce existing datasets which can be used for TIR pedestrian tracking evaluation in Section~\ref{existingdataset}. Then, we discuss the progress of TIR pedestrian tracking methods in Section~\ref{existingtrackers}.

\subsection{TIR Pedestrian Tracking Datasets}
\label{existingdataset}
There is not a standard and specialized TIR pedestrian tracking dataset but several datasets can be used to simply test the TIR pedestrian tracker.

\vspace{3mm}
\noindent{\textbf{OSU Color-Thermal.}} The original purpose of this dataset~\cite{OSU} is to do object detection using RGB video and thermal video. The dataset contains $6$ TIR pedestrian videos captured from a fixed thermal sensor Raytheon PalmIR 250D. These videos have a low resolution at $320\times 240$ pixels and their backgrounds are static.

\vspace{3mm}
\noindent{\textbf{Terravic Motion IR.}} This dataset~\cite{OTCBVS} is designed for a detection and tracking task in thermal video. It has $18$ TIR sequences {captured from a Raytheon L-3 Thermal-Eye 2000AS thermal camera}. Among these sequences, $11$ pedestrian sequences are suitable for tracking task. These sequences are all in the same outdoor scene with a static background and a low resolution at $320 \times 240$ pixels.

\vspace{3mm}
\noindent{\textbf{PDT-ATV.}} PDT-ATV~\cite{PDT-ATV} is a TIR pedestrian tracking and detection dataset which is captured from a simulated unmanned aerial vehicle (UAV). The dataset contains $8$ sequences with the same resolution at $324\times 256$ pixels. The dataset provides ground-truths of the object but does not annotate attributes of each sequence.

\vspace{3mm}
\noindent{\textbf{BU-TIV.}} BU-TIV~\cite{TIV} is a large-scale dataset for several visual analysis tasks in TIR videos. It contains $5$ TIR pedestrian videos that can be used for tracking task. The resolution of these videos ranges from $512\times 512$ to $1024\times 640$. The dataset provides annotations of the object but does not offer any evaluation codes for tracking task.

Table~\ref{comd} compares the above-mentioned datasets with our dataset.
Although these datasets can be used in TIR pedestrian tracking to test a tracker, they are not suitable for fair comparison and evaluation.
In this paper, in order to fairly compare and evaluate a tracker, we collect a large-scale TIR pedestrian tracking benchmark dataset with $60$  annotated sequences.

\subsection{TIR Pedestrian Tracking Methods}
\label{existingtrackers}
In the past decade, numerous TIR pedestrian tracking methods have been proposed to solve various challenges. Similar to visual object tracking~\cite{dong2017occlusion,liu2019msst,zhang2015object,HSSNet,he2017robust,ma2017saliency,Multi-viewCF,he2016connected,ma2018visual} and grayscale-thermal tracking~\cite{li2017grayscale}, there are two categories of TIR pedestrian trackers: generative and discriminative. Generative TIR pedestrian trackers focus on the modeling of the pedestrian's appearance {at current frame} and search for the most similar candidates in next frame. Some representative methods are template matching~\cite{dai2007pedestrian,jungling2010pedestrian}, sparse representation~\cite{li2014robust,liang2016local}. Unlike the generative methods, discriminative TIR pedestrian tracking methods cast the tracking problem as a binary classification problem, which distinguishes the object target from its backgrounds. Thanks to advances in machine learning, a series of classifiers can be used in TIR pedestrian tracking, such as random forest~\cite{ko2013human}, mean shift~\cite{xu2005pedestrian}, and support vector machines (SVM)~\cite{Wang2006Target}, linear discriminant analysis~\cite{wen2018robust}. Here, we discuss three components of these methods: feature extractor, motion model, and observation model.

\captionsetup[subfigure]{labelformat=empty, justification=centering}
\begin{figure*}[htbp]
    \centering
    \includegraphics[width=1\textwidth]{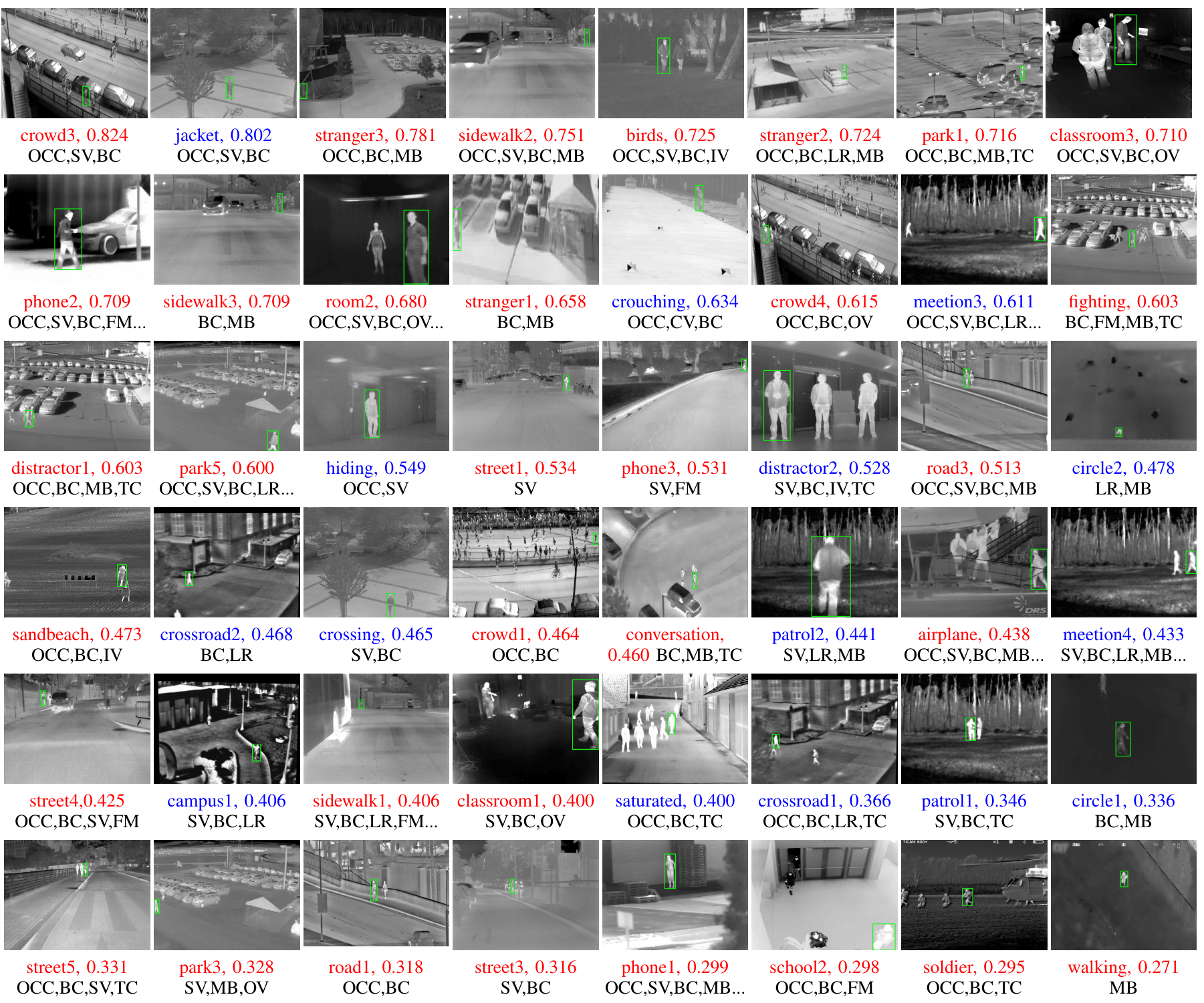}\\
   \caption{Some annotated sequences of the proposed dataset. The target's ground-truth bounding box in the first frame is shown. The sequences are ranked according to their difficulty degree, which is the average TRE score of the tested trackers on each sequence (see the \textbf{supplemental material}).
   The sequence name, difficulty degree, and corresponding attributes are shown below the image, respectively.
   {\color{blue}Blue} font denotes old sequences used in the previous studies, while {\color{red} red} font denotes newly collected sequence. The signs, such as, OCC, SV and BC, indicates the attributes of one sequence. More information of these signs can be found in Table \ref{attributes}.}
\label{firstframeBB}
\end{figure*}

\vspace{3mm}
\noindent{\textbf{Feature Extractor.}} Obtaining a powerful representation of the object target is a crucial step in TIR pedestrian tracking. In contrast to visual pedestrian tracking that often uses the color feature, the intensity feature is widely used in TIR pedestrian tracking due to a basic assumption that the object target is warmer than its background in thermal images. However, a tracker only using the intensity feature often fails when two similar pedestrians are occluded by each other. In order to get a more discriminative feature representation, several TIR pedestrian trackers exploit multiple features in a fusion strategy.
For example, Wang et al.~\cite{wang2010modified} {combine} the intensity and edge cues via an adaptive multi-cue integration scheme.
In~\cite{wang2012pedestrian}, the authors also fuse the  intensity and edge information using a relative discriminative coefficient.
In~\cite{ko2013human}, the authors coalesce local intensity distribution (LID) and oriented center symmetric local binary patterns (OCS-LBP) to represent the TIR pedestrian.
Furthermore, some other features are often used in TIR pedestrian tracking, such as regions of interest (ROI) histograms~\cite{li2010real}, speeded up robust features (SURF)~\cite{jungling2010pedestrian}, and the histogram of oriented
gradients (HOG)~\cite{kim2015pedestrian}.

\vspace{3mm}
\noindent{\textbf{Motion Model.}} The ability to quickly and accurately generate a series of target candidates is another important component in TIR pedestrian tracking. Usually, two kinds of methods are used to generate target candidates which are the probabilistic estimation and the exhaustive search. For probabilistic estimation methods, Kalman filter and particle filter are often exploited in TIR pedestrian tracking. For instance, Xu et al.~\cite{xu2005pedestrian} combine the Kalman filter and the mean shift to track pedestrians with a vehicle-mounted thermal camera. Wang et al.~\cite{wang2010modified} propose a multiple features fusion TIR pedestrian tracker based on the particle filter framework.
Several other TIR pedestrian tracking methods~\cite{li2010real,wang2012pedestrian,ko2013human} also use the particle filter.  For exhaustive search methods, two kinds of methods are often used in visual tracking including sliding window~\cite{wang2015understanding} and radius sliding window~\cite{hare2016struck}, which are also suitable for TIR pedestrian tracking.

\vspace{3mm}
\noindent{\textbf{Observation Model.}} In the discriminative TIR pedestrian tracking method, observation models are used to obtain the tracked target from a series of candidates. Usually, we train a binary classifier to complete this task. In the past decade, numerous classification methods have been used. For example, Xu et al.~\cite{xu2005pedestrian} use the mean shift to get the final tracked target from a series of target candidates generated from the Kalman filter.  Ko et al.~\cite{ko2013human} {exploit a} random forest to obtain the tracked target from candidates. Some other classification methods {such as SVM~\cite{Wang2006Target}, boosting~\cite{grabner2006real}, and multiple instance learning (MIL)~\cite{babenko2009visual} are also suitable for TIR pedestrian tracking}.

These three components of the TIR pedestrian tracker affect the tracking performance in different ways. How does each component affect the final tracking results? In this paper, {we report the results of several validation experiments to answer this question.}

\section{TIR Pedestrian Tracking Benchmark}
\label{benchmark}
In this section, we first introduce our TIR pedestrian tracking dataset in Section~\ref{dataset}, and then, present the evaluation methodology for the TIR pedestrian tracker in Section~\ref{em}. These two parts constitute our TIR pedestrian tracking benchmark.

\subsection{Dataset}
\label{dataset}
In the past decade, several datasets~\cite{OSU,OTCBVS,PDT-ATV,TIV} have been used to evaluate the TIR pedestrian tracker. However, these datasets do not provide uniform ground-truths for the fair evaluation.  To implement fair performance evaluation and comparison for TIR pedestrian trackers, we collect $60$ thermal sequences and then annotate them  manually.
These sequences come from different devices, scenes, and shooting times. For each shooting property, we collect a series of corresponding thermal videos. More basic properties can be found in Table~\ref{generalproperty}.
These different properties ensure the diversity of the dataset.
\begin{table}[htbp]
  \centering
  \caption{Basic properties of the shooting videos and the corresponding video number. }
    \rowcolors{2}{gray!25}{white}
    \begin{tabular}{l|p{5.8cm}}
    \hline
    \rowcolor{gray!50}
    Property Name   & Property value: video number \\
    \hline
    Device Categories & Surveillance camera: 29; Hand-held camera: 20; Vehicle-mounted camera: 8; Drone camera: 3 \\
    Scene Types & Outdoor: 52; Indoor: 8 \\
    Shooting Time & Night: 42;  Day: 18 \\
    Camera Motion & Static: 43; Dynamic: 17 \\
    Camera Views & Down: 31; Level: 29 \\
    Object Distance & Far: 21; Middle: 23; Near: 16 \\
    Object Size & Big: 12; Middle: 34; Small: 14 \\
    \hline
    \end{tabular}%
  \label{generalproperty}%
\end{table}%

\noindent{\textbf{Sources.}} Our datasets are collected from existing commonly used thermal sequences and video websites. Four thermal sequences are obtained from OSU Color-Thermal dataset~\cite{OSU}.  Six sequences are adopted from Terravic Motion IR~\cite{OTCBVS} and nine sequences are chosen  from BU-TIV~\cite{TIV}. Two sequences come from LITIV2012~\cite{litiv2012} and five sequences come from detection dataset CVC-09~\cite{CVC-09} and CVC-14~\cite{CVC-14}. In addition, we selected seven pedestrian videos from the TIR object tracking benchmark: VOT-TIR2016~\cite{VOT-TIR2016}. The rest of sequences are collected from the Internet including INO dataset~\cite{INO} and YouTube~\cite{Youtube}.

\vspace{3mm}
\noindent{\textbf{Annotations.}} An external rectangle bounding box of the target is used as its ground-truth. The first frame annotations of several sequences are shown in Fig.~\ref{firstframeBB}. The left corner point, width, and height of the target's bounding box are recorded to represent the ground-truth.

\vspace{3mm}
\noindent{\textbf{Attributes of a Sequence.}} Evaluating a TIR pedestrian tacker is usually difficult because several attributes can affect its performance.
In order to better evaluate the strength and weakness of a TIR pedestrian tracker, we summarize nine common attributes in the sequences, as shown in Table~\ref{attributes}.
For each attribute, we construct a corresponding subset for evaluation. The performance of a tracker on an attribute subset shows its ability for handling the corresponding challenge.
The attributes distribution of the entire dataset and an attribute subset are shown in Fig.~\ref{attrdis}.
The other subset's attributes distribution is shown in the \textbf{supplemental material}.
We can see that the background clutter has a high ratio because pedestrians often have a similar texture and intensity in thermal images. In addition, occlusion and scale variation also often occurs in real-world scenarios. The intensity variation subset only includes three sequences as pedestrians tend to have an invariable temperature over a short time period.

\begin{table}[htbp]
  \centering
  \caption{Attributes of a thermal sequence. }
    \rowcolors{2}{gray!25}{white}
    \begin{tabular}{l|p{7cm}}
    \hline
    \rowcolor{gray!50}
    Challenge   & Description \\
    \hline
    TC   & Thermal crossover: Two targets with similar intensity cross each other. \\
    IV   & Intensity variation: the intensity of the target region has changed due to the temperature variation of the target. \\
    OCC & Occlusion: the target is partially or fully occluded. \\
    SV   & Scale variation: the ratios of the target's size in the first frame and current frame is out of the range [1/2, 2].  \\
    BC   & Background clutter: the background near the target has similar texture or intensity. \\
    LR   & Low resolution: the target's size is lower than $600$ pixels.  \\
    FM   & Fast motion: the distances of target in the consecutive frame are larger than 20 pixels. \\
    MB   & Motion blur: the target region is blurred due to target or camera motion. \\
    OV   & Out-of-view: the target is partially out of image region. \\

    \hline
    \end{tabular}%
  \label{attributes}%
\end{table}%

\begin{figure}[htbp]
  \centering
   \subfloat[]{\includegraphics[width=0.24\textwidth]{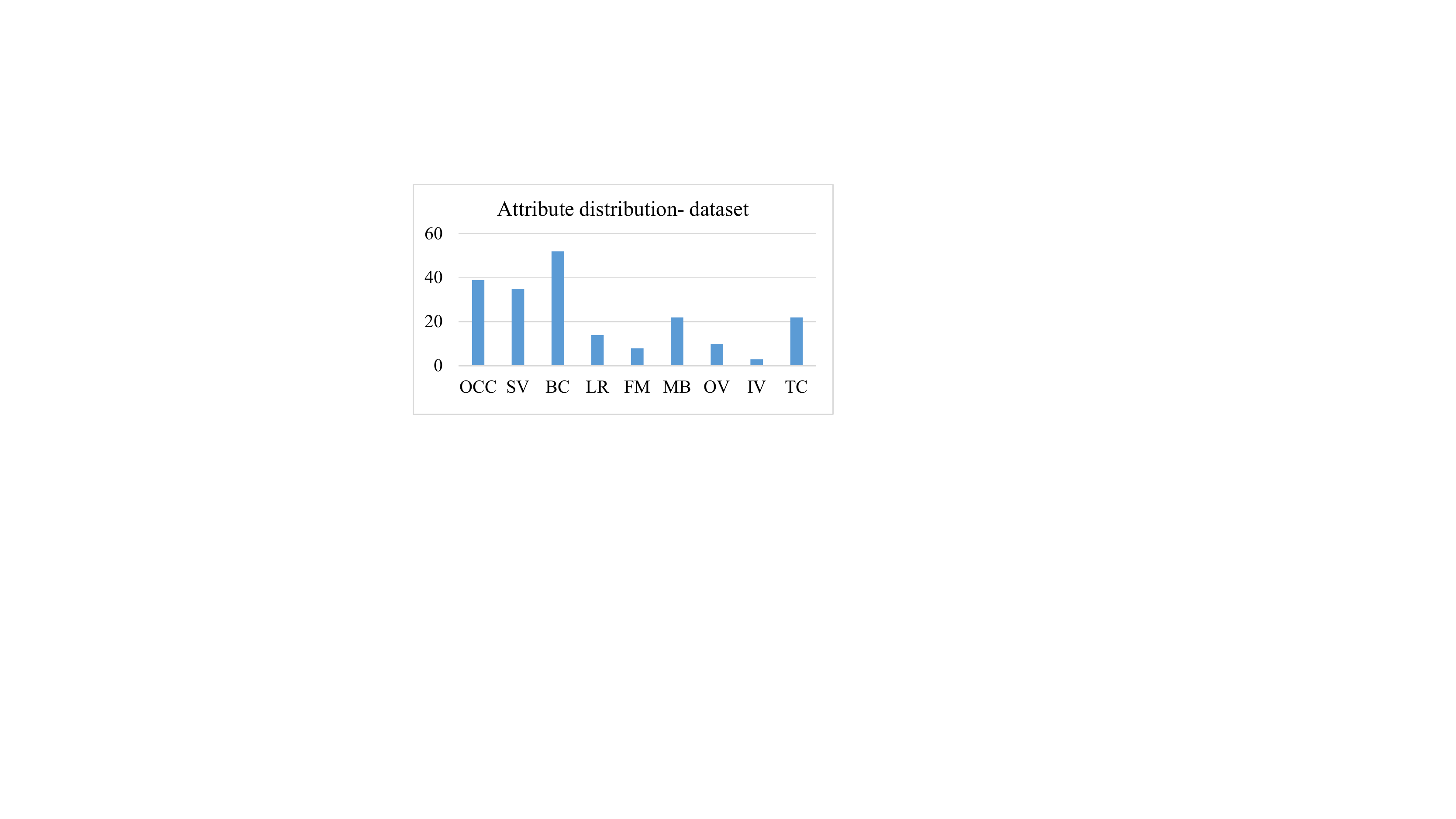}}
   \subfloat[]{\includegraphics[width=0.24\textwidth]{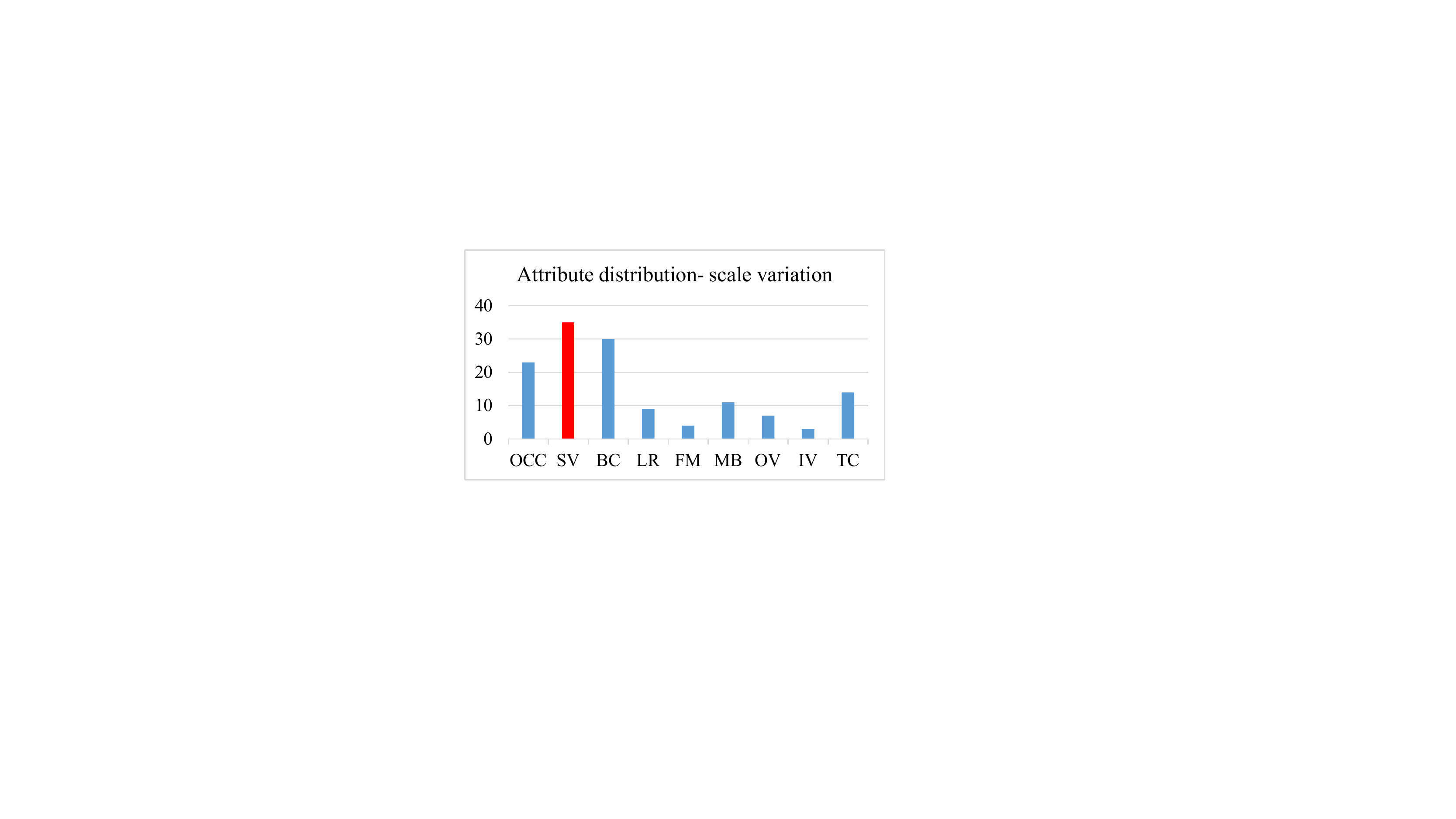}}\vspace{-0.1in} \\
  \caption{The attributes distribution of the entire dataset and a scale variation subset.}
  \label{attrdis}
\end{figure}

\subsection{Evaluation Methodology}
\label{em}
There are two commonly used evaluation metrics in visual object tracking: center location error (CLE) and overlap ratio. In this paper, we also adopt these two metrics to evaluate TIR pedestrian trackers. Following OTB~\cite{OTB50}, we use the precision and success rate for quantitative analysis.

\vspace{3mm}
\noindent{\textbf{Precision Plot.}} CLE is an average Euclidean distance between the tracked target and target's ground-truth. If a CLE is within a given threshold (20 pixels), we say the tracking is successful at this frame. The precision plot measures the percentage of the successful frames in the entire dataset.

\vspace{3mm}
\noindent{\textbf{Success Plot.}} Overlap score measures the overlap ratio between the bounding box area of the tracked target and ground-truth. If a score is larger than a given threshold, we say the tracking is successful at this frame. The success plot shows the ratios of successful frames at the threshold varying from $0$ to $1$. The area under the curve (AUC) of each success plot is exploited to rank trackers.

\vspace{3mm}
\noindent{\textbf{Robustness Evaluation.}} Usually, a tracker is sensitive to initialization, and one-pass evaluation (OPE) does not provide the robustness evaluation. In order to measure a tracker's robustness to different initializations, we use the temporal robustness evaluation (TRE) and the spatial robustness evaluation (SRE). TRE runs 20 times with different initial frames in a sequence and then calculates the average precision and success rate. SRE runs $12$ times with a different initial bounding box by shifting or scaling.

\vspace{3mm}
\noindent{\textbf{Speed Evaluation.}} Speed is the other important aspect of a tracker. We run all trackers on the same hardware device and calculate their average frames per second (FPS) in the entire dataset.

\section{Evaluation Experiments}
\label{exp1}
In this section, we discuss the implementation of large-scale fair evaluation experiments on the proposed benchmark. First, an overall performance evaluation of nine trackers is reported in Section~\ref{OP}. Second, we discuss our evaluation of the performance of these trackers on the attribute subset in Section~\ref{ABE}. Third, a speed comparison experiment is presented in Section~\ref{SC}.

\subsection{Overall Performance Evaluation}
\label{OP}
\noindent{\textbf{Evaluated Trackers.}} Nine publicly available trackers are evaluated on our benchmark. {Since existing TIR pedestrian trackers} have no publicly available codes, we chose some commonly used visual trackers and TIR trackers for evaluation. These trackers can be used for TIR pedestrian tracking and can be generally divided into four categories:
\begin{itemize}
  \item Three correlation filter based trackers { including} kernelized correlation filters (KCF~\cite{KCF}), scale correlation filters (DSST~\cite{DSST}), and spatially regularized correlation filters (SRDCF~\cite{SRDCF}).
  \item Two deep learning based trackers including HDT~\cite{HDT} and MCFTS~\cite{MCFTS}.
  \item Two regression based trackers which are ridge regression (RR~\cite{wang2015understanding}) and gaussian regression (TGPR~\cite{TGPR}).
  \item Two other trackers which are SVM~\cite{wang2015understanding} and sparse representation (L1APG~\cite{L1APG}).
\end{itemize}
Most of these trackers have state-of-the-art performance in visual tracking. For all of these trackers, we use the original parameters of them for evaluation.

\begin{figure}
  \centering
   \subfloat[]{\includegraphics[width=0.24\textwidth]{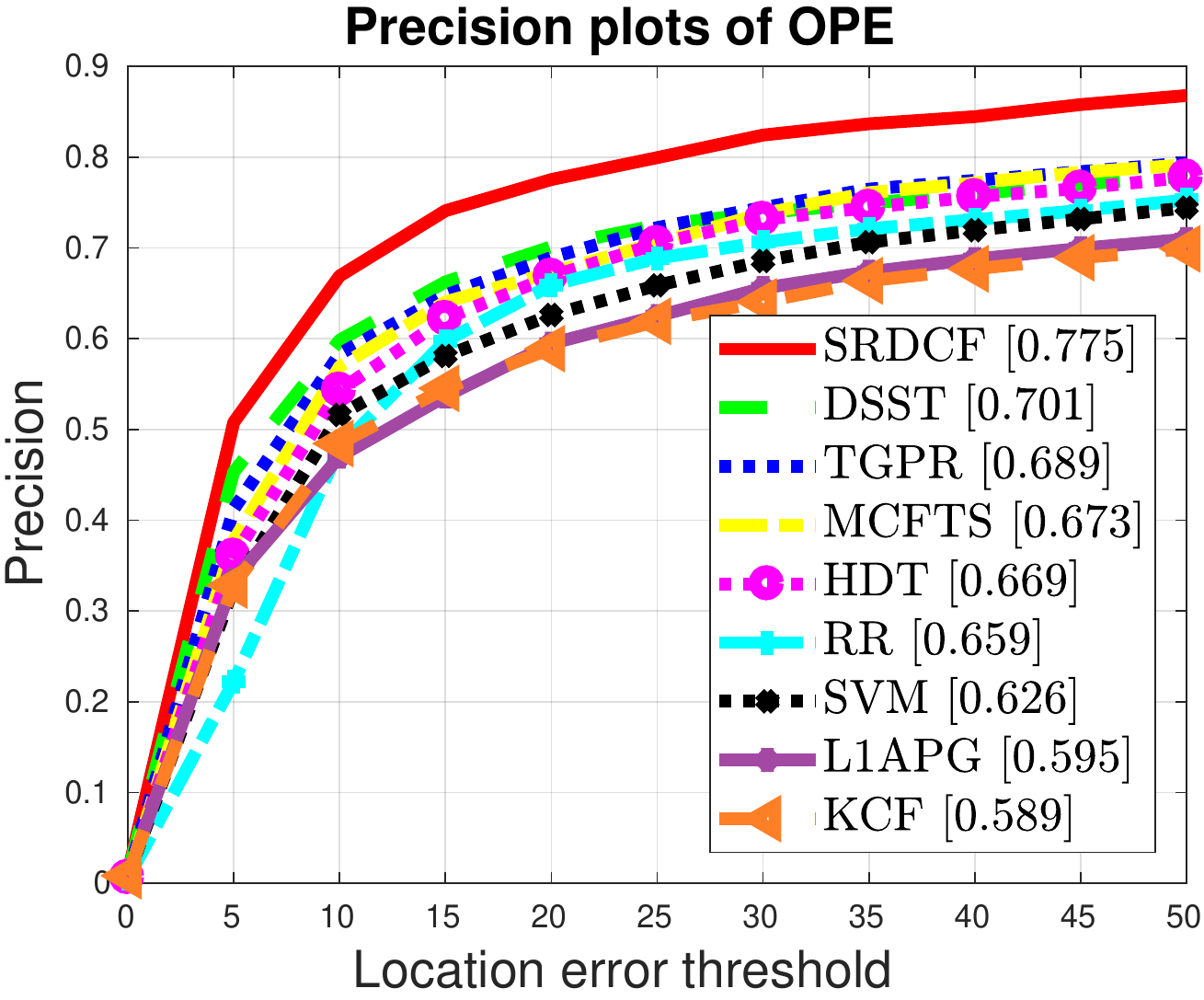}
                \includegraphics[width=0.24\textwidth]{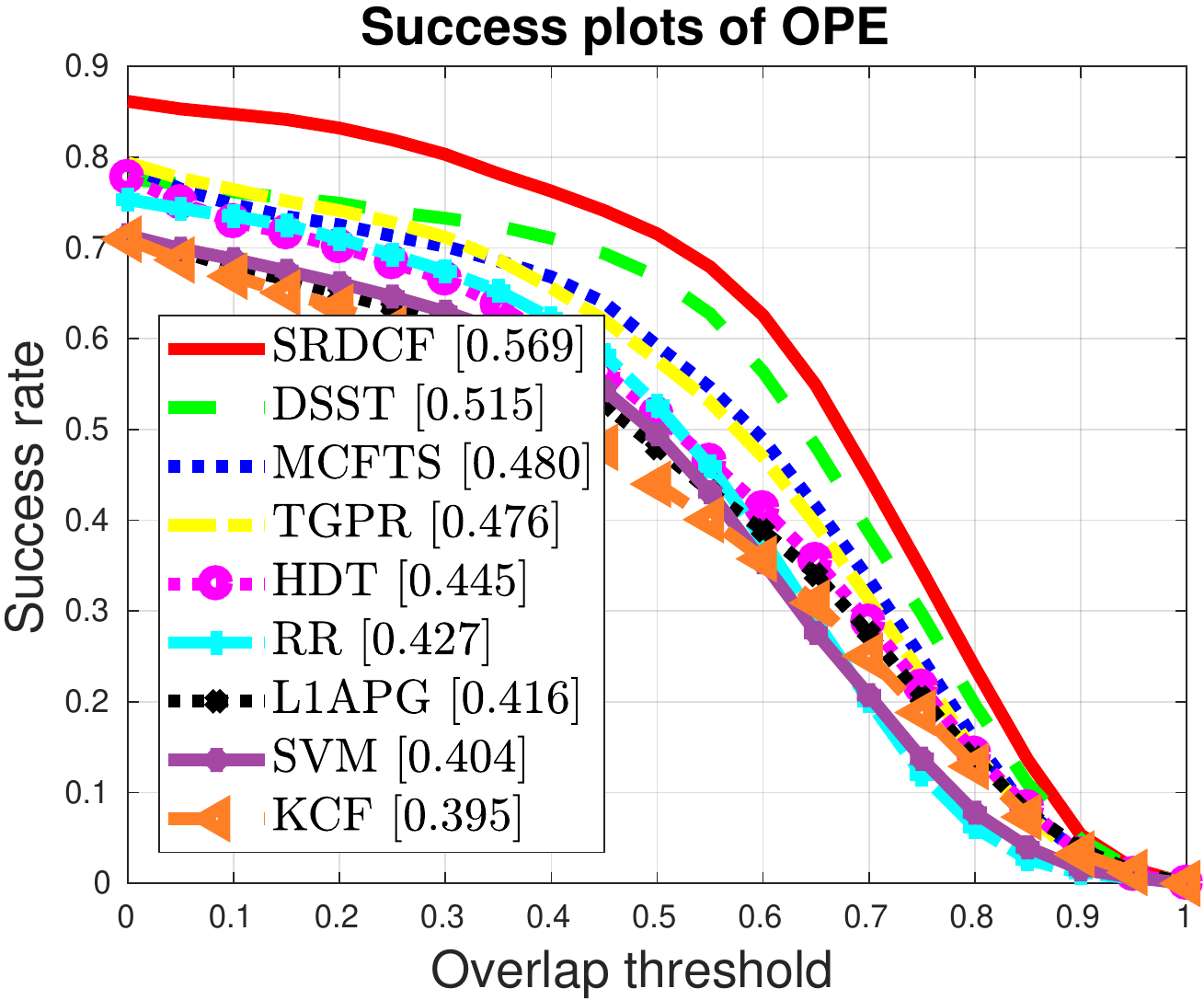}}\vspace{-0.15in} \\
    \subfloat[]{\includegraphics[width=0.24\textwidth]{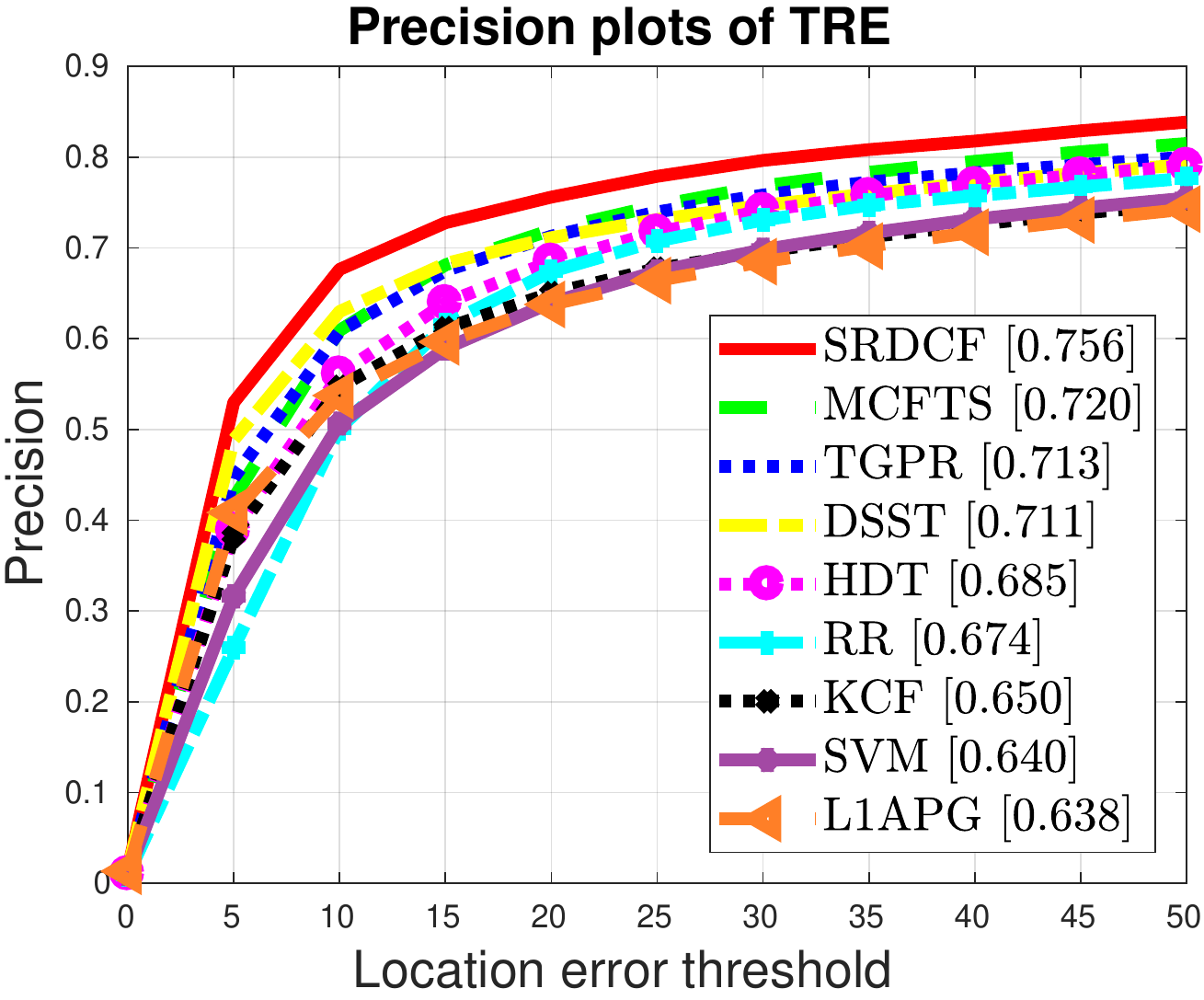}
                \includegraphics[width=0.24\textwidth]{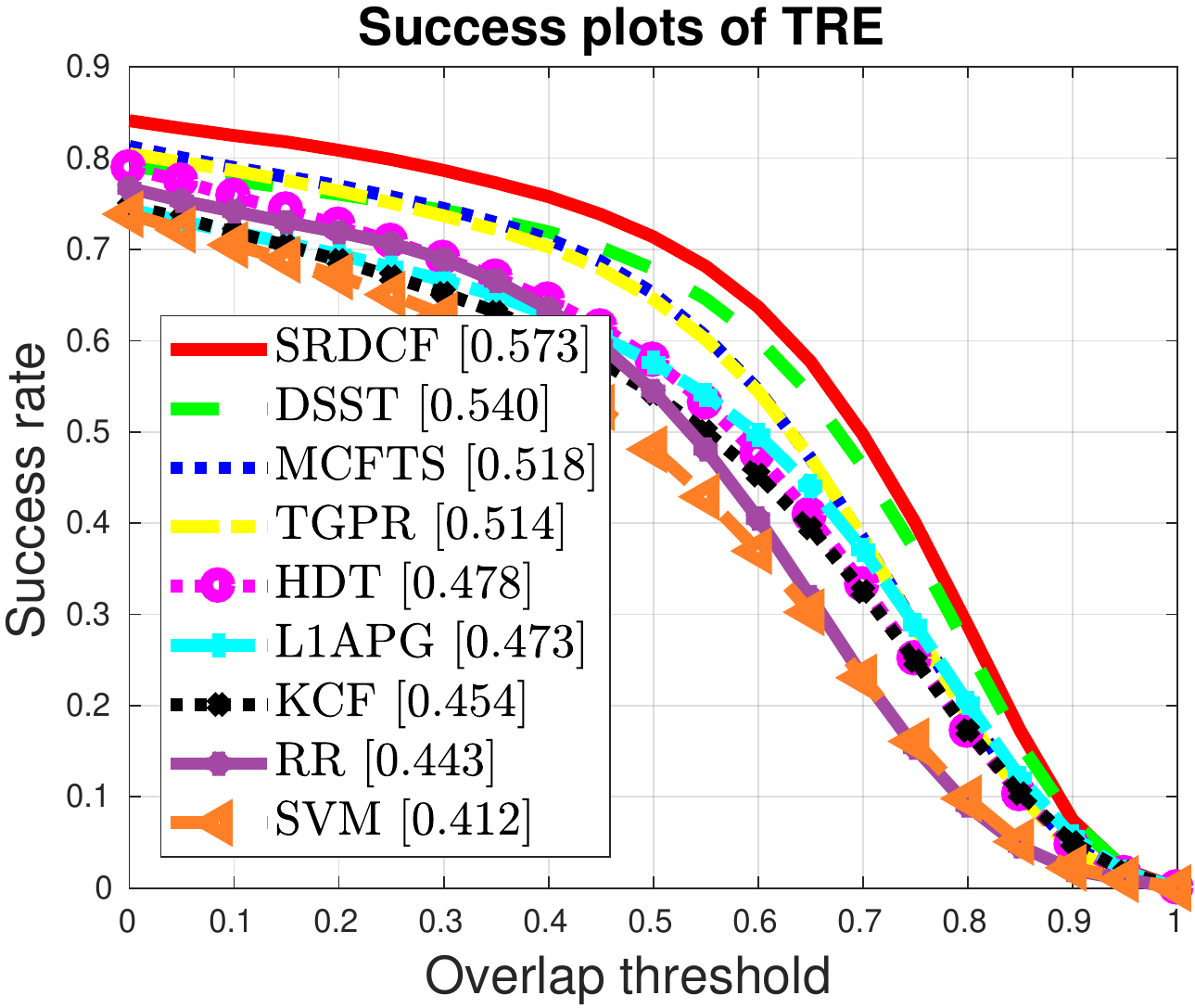}}\vspace{-0.15in} \\
      \subfloat[]{\includegraphics[width=0.24\textwidth]{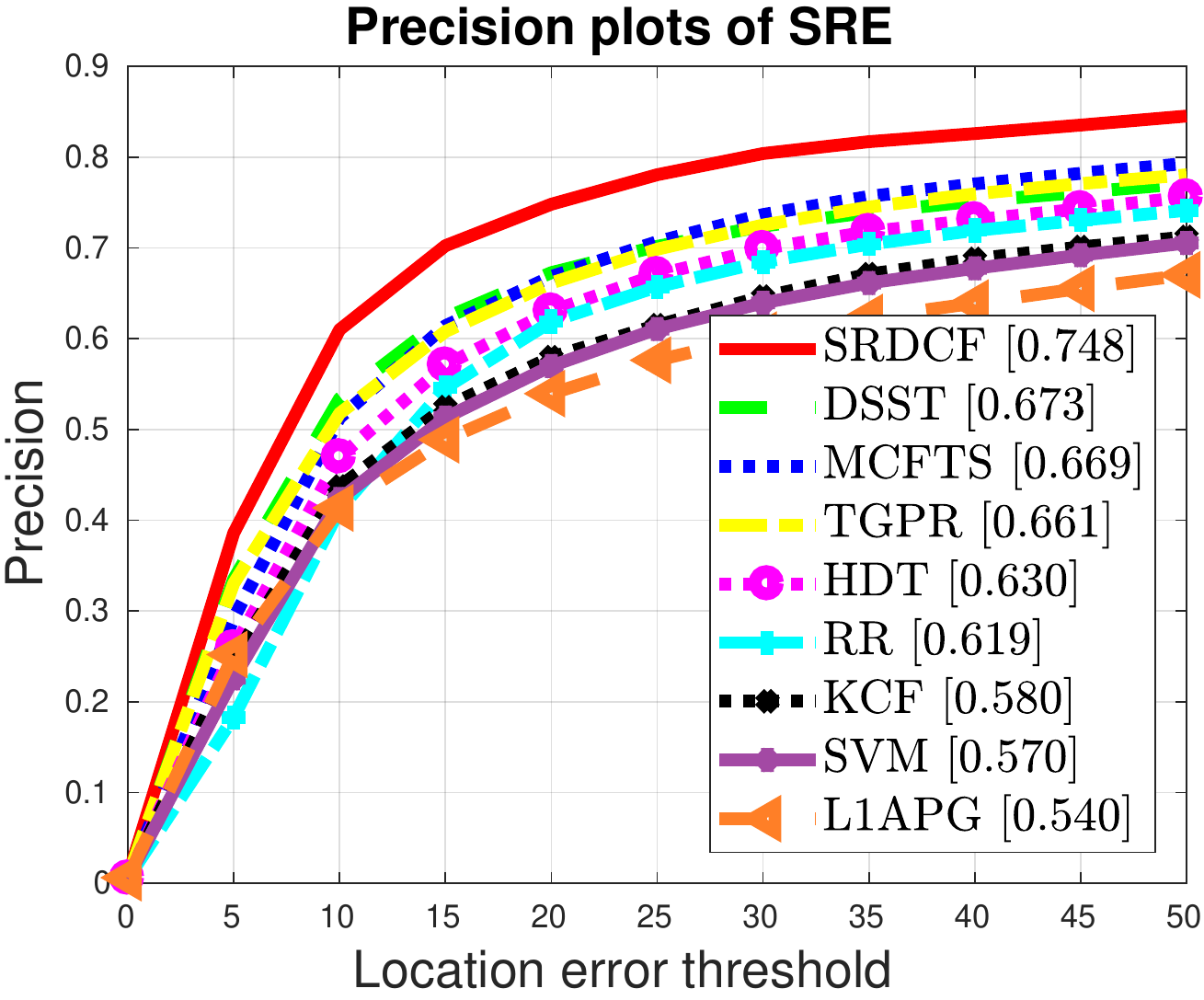}
                   \includegraphics[width=0.24\textwidth]{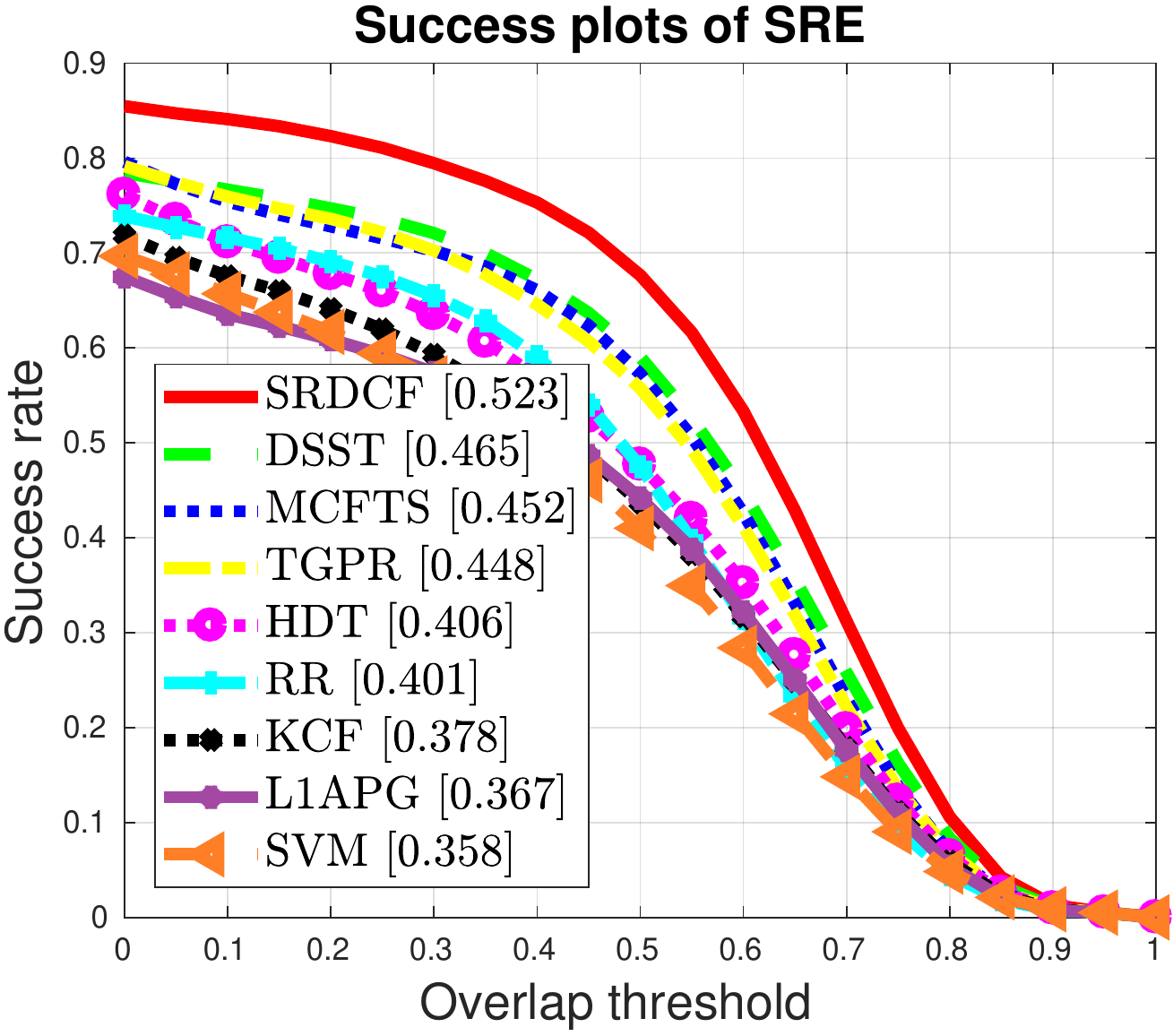}}
  \caption{The overall performance of the trackers using OPE, TRE, and SRE plots. The performance score is shown in the legend of the figures. The trackers are ranked in descending order according to their performance. }
  \label{Overallperformance}
\end{figure}

\begin{figure*}
  \centering
  \includegraphics[width=1\textwidth]{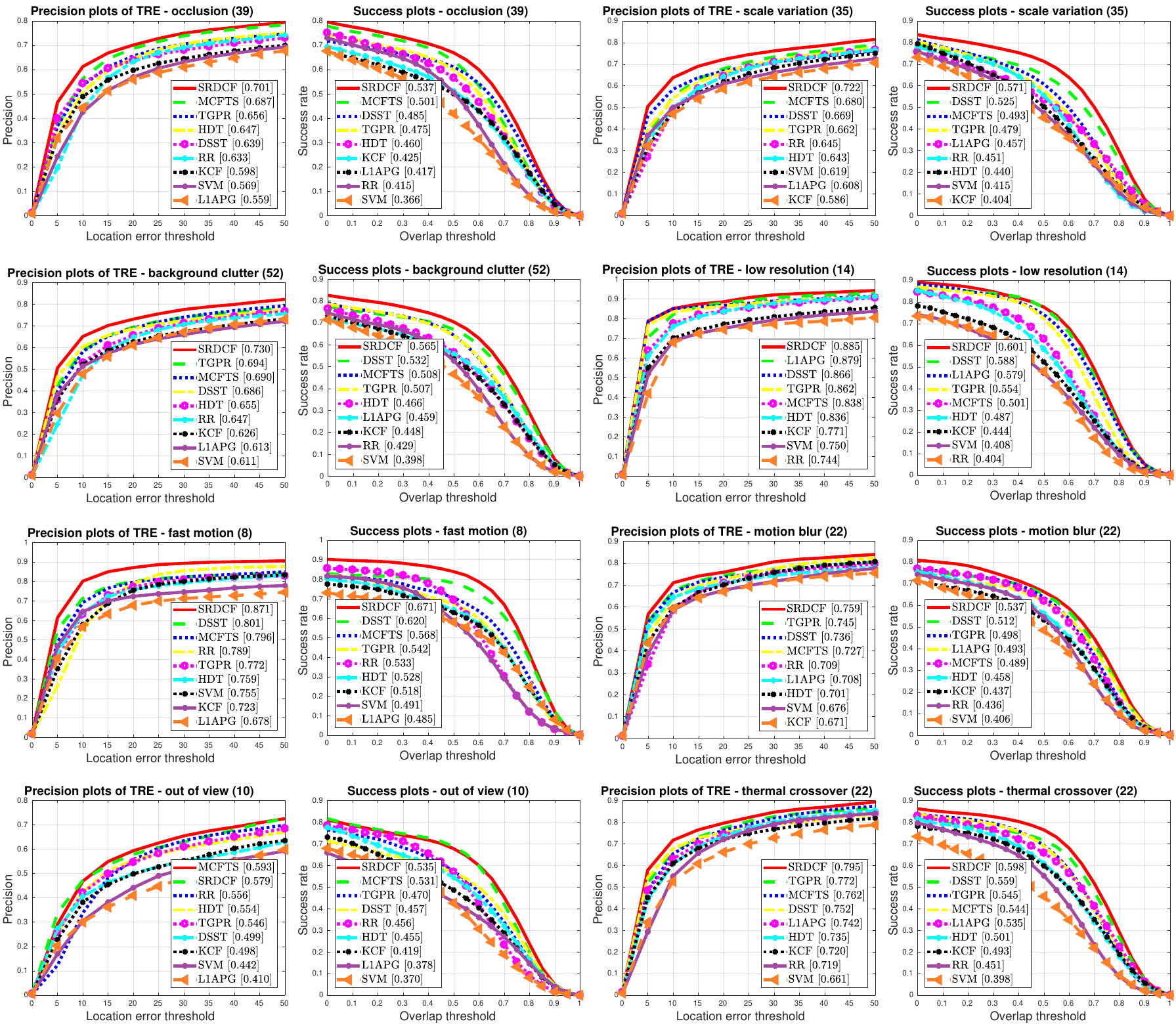}\\
  \caption{TRE plots on eight attribute subsets. The title of each sub-figure consists of the attribute name and corresponding sequences number. }
  \label{attributesperformance}
\end{figure*}

\vspace{3mm}
\noindent{\textbf{Evaluation Results.}} Precision plots and success plots are used to show the overall performance of the tracker, and the results are shown in Fig.~\ref{Overallperformance}. OPE is used to show the precision and success rate of the tracker, while TRE and SRE are used to show the robustness of the tracker.

\vspace{3mm}
\noindent{\textbf{Analysis.}} The trackers are ranked according to their performance score. From Fig.~\ref{Overallperformance}, we can see that the ranking of the trackers is slightly different in the precision plots and success plots. This is because the evaluation metrics are different in these two plots. Precision  is ranked  according to the percentage of successful frames where center location error is within a fixed threshold (20 pixels), while the success rate is ranked according to AUC scores. We suggest that the success plots are more accurate than the precision plot for ranking. Therefore, in the following, we mainly use the success plots to analyze the tracker's ranking and use the precision plots as an auxiliary.

The tracker's performance in TRE is higher than in OPE because the trackers tend to perform well in a shorter sequence. The OPE result is just one trial of TRE. On the contrary, the tracker's performance in SRE is lower than in OPE since the initialization errors tend to cause the drift of the trackers.

In the success plots, we can see that SRDCF ranks the first. That means its success rate and robustness are the best in all of these trackers. It is noteworthy that SRDCF is much better than others. In OPE and SRE, the performance of SRDCF is higher than the second best DSST by more than $5\%$. In TRE, it is also higher than DSST by about $3\%$. We suggest that its superior performance benefits from the consideration of more background information in tracking. On the other hand, it is easy to see that the performance of DSST is higher than KCF by more than $10\%$. This shows that the scale estimation strategy is very important in TIR pedestrian tracking.

Deep feature based TIR tracker MCFTS achieves a promising result despite the fact that the deep features are learned from RGB images. We suggest that the deep feature based trackers have the potential to achieve better performance if there are enough thermal images for training. On the other hand, MCFTS just uses a simple combination method between the deep network and KCF. We believe that {a deeper level of} integration between them can obtain more robust tracking results.  In addition, we can see that the performance of Gaussian regression tracker TGPR is higher than ridge regression tracker RR by about $5\%$. This is a prominent improvement. Therefore, we think a more complex kernel function is better than a simple one in the regression-based TIR pedestrian tracker.

\begin{table*}[htbp]
  \centering
  \caption{Speed comparison of the trackers. }
    \rowcolors{2}{gray!25}{white}
    \begin{tabular}{c|c|c|c|c|c|c|c|c|c}
    \hline
    \rowcolor{gray!50}
   ~ &KCF \cite{KCF}   & DSST \cite{DSST} & SVM \cite{wang2015understanding} & SRDCF \cite{SRDCF} & HDT \cite{HDT} & RR \cite{wang2015understanding} & MCFTS \cite{MCFTS} &L1APG \cite{L1APG} &TGPR \cite{TGPR} \\
    \hline
   FPS & 393.40 & 96.30 & 13.40 & 12.29 & 10.60 & 6.64 & 4.73 & 3.66 & 1.77  \\
    \hline
    \end{tabular}%
  \label{speed}%
\end{table*}%

\subsection{Attribute-based Evaluation}
\label{ABE}
\noindent{\textbf{Results.}} In order to understand the tracker's performance for different challenges, we evaluate nine trackers on the attribute subsets.
The results are shown in Fig.~\ref{attributesperformance}. Here, we just show the TRE plots on these subsets due to space limitation. OPE and SRE plots are presented in the \textbf{supplemental material}.

\vspace{3mm}
\noindent{\textbf{Analysis.}} The tracker's performance on an attribute subset shows its ability to handle this challenge.  As shown in Fig.~\ref{attributesperformance}, we can see that SRDCF achieves the best performance on almost all attribute subsets because it effectively solves the boundary effect brought from the cyclic shift.  MCFTS achieves better performance on the occlusion and out-of-view subsets than DSST, despite DSST having a higher overall performance than MCFTS (see TRE of Fig.~\ref{Overallperformance}). That shows that deep learning based trackers have a promising performance in TIR pedestrian tracking. KCF and HDT have the worst performance on the scale variation subset since they lack a scale estimation strategy, while DSST obtains the second-best performance on the scale variation subset and the entire dataset since it deals with the scale variation. This demonstrates that the scale estimation strategy is very useful for improving the tracking performance. On the fast motion subset, we can see that RR achieves a much higher success rate than its overall performance (see Fig.~\ref{Overallperformance}). We suggest that the stochastic particle filter search framework plays a major role.  These results help us understand the strengths and weaknesses of the trackers.


\subsection{Speed Comparison}
\label{SC}
We carry out experiments on a same PC with an Intel I7-6700K CPU, 32G RAM, and a GeForce GTX 1080 GPU card. The speed of the tracker is the average FPS on the TRE results. The comparison of the tracker's speed is shown in Table~\ref{speed}.

\vspace{3mm}
\noindent{\textbf{Analysis.}} Table~\ref{speed} shows that KCF has a high speed while DSST also exceeds real-time speed. The high efficiency of these two trackers benefits from the computation in the Fourier domain. In addition, SRDCF obtains half of the real-time speed when it achieves the best precision and success rate.  Two deep learning based trackers, HDT and MCFTS, just obtain a low frame rate due to the high cost of deep feature extraction.
Several classifier based trackers, such as SVM and RR also have a low speed because they are based on a time-consuming particle filter search framework.

\section{Validation Experiments}
\label{exp2}
In this section, we report three comparison experiments that we conduct to validate each component's contribution to TIR pedestrian tracking performance. First, we carry out a comparison experiment on two baseline trackers with seven different features, which is discussed in Section~\ref{FE}. Then, a comparison experiment on a baseline tracker with three different motion models is conducted, as is discussed in Section~\ref{MM}. Finally, in Section~\ref{OM}, we discuss our comparison of several different observation models to validate how they affect tracking performance.

\subsection{Feature Extractor.}
\label{FE}
\begin{figure}
  \centering
   \subfloat[{(a) Ridge regression}]{\label{RR}\includegraphics[width=0.24\textwidth]{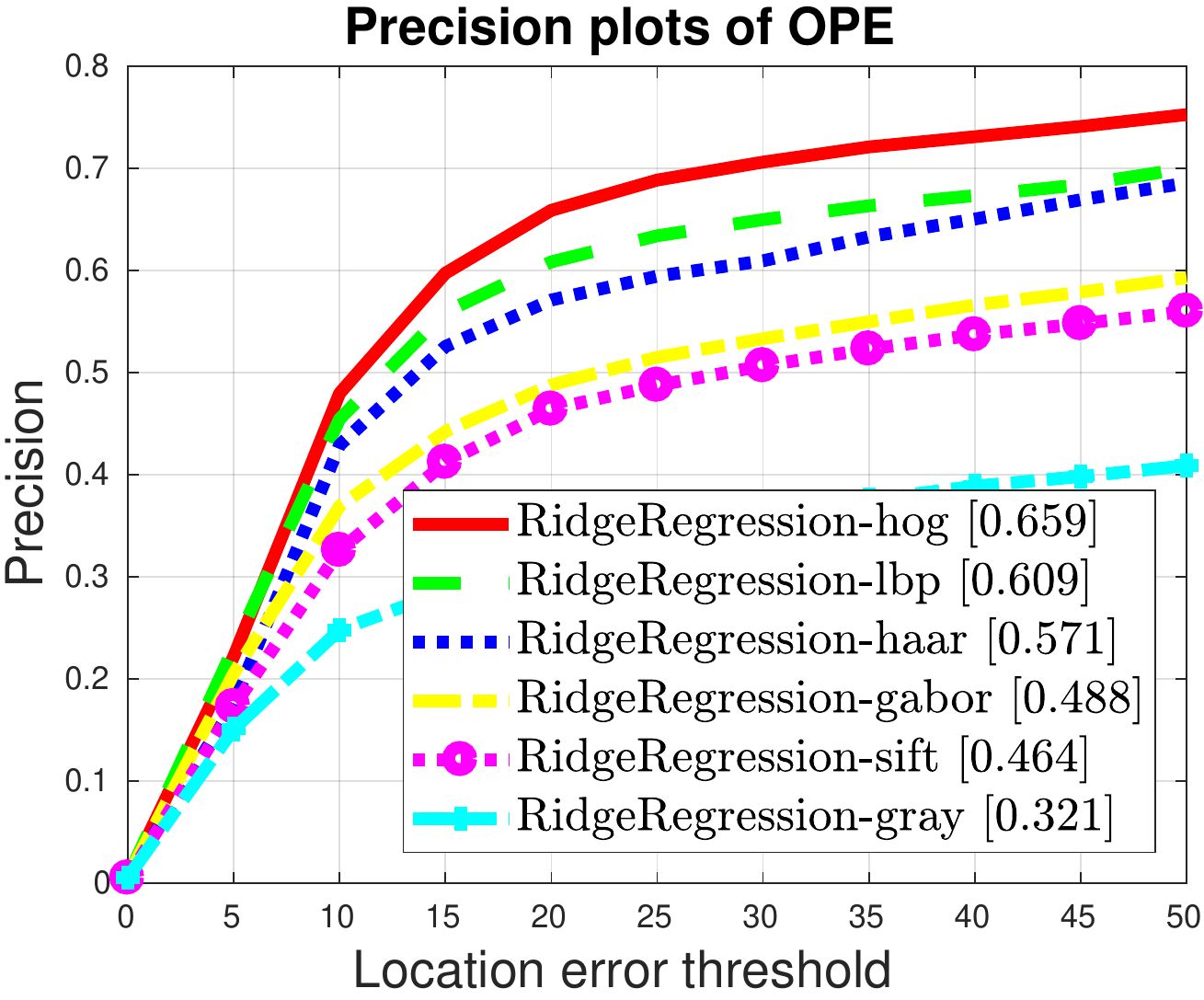}
               \includegraphics[width=0.24\textwidth]{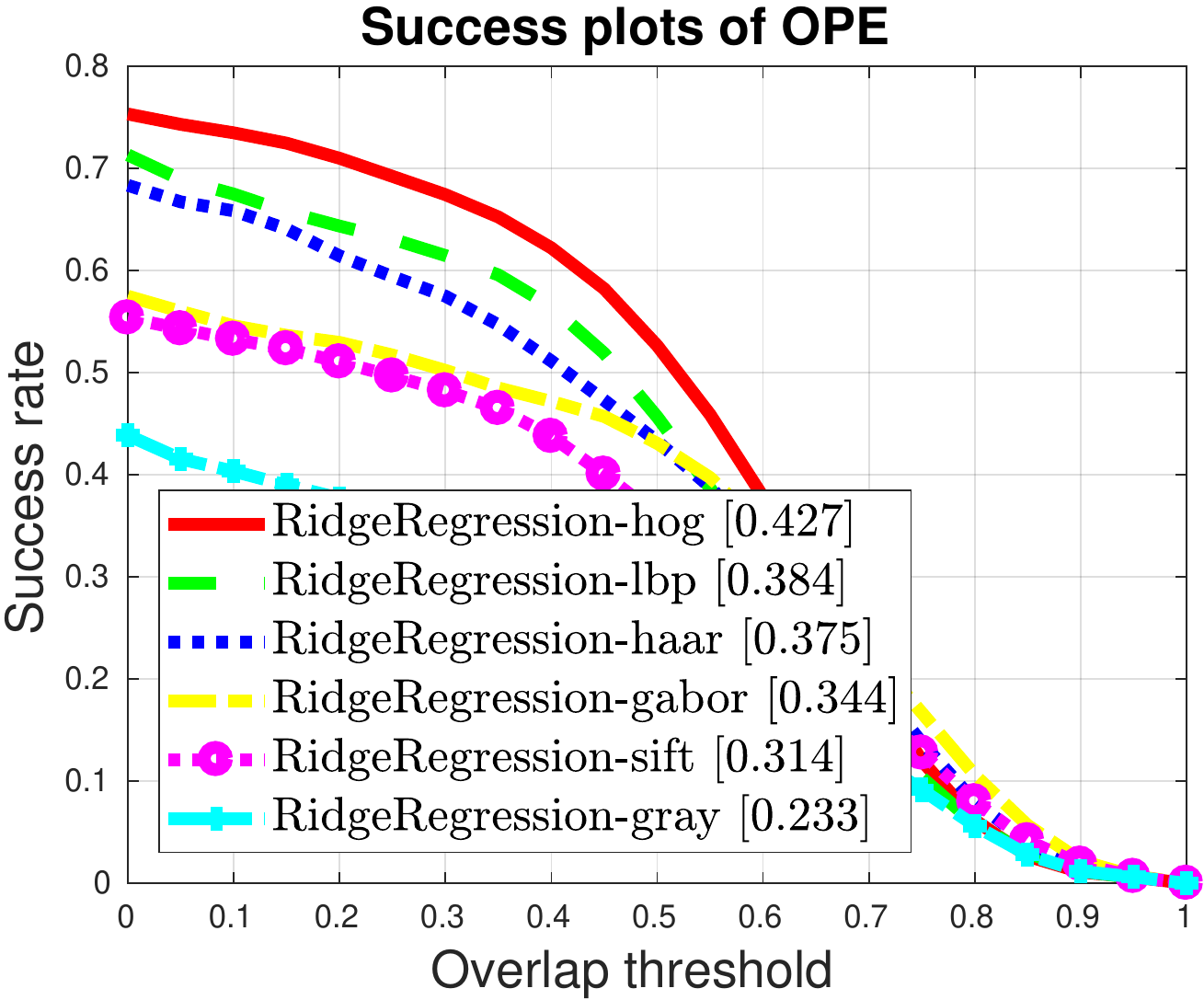}} \vspace{-0.1in}\\
   \subfloat[{(b) KCF}]{\label{KCF}\includegraphics[width=0.24\textwidth]{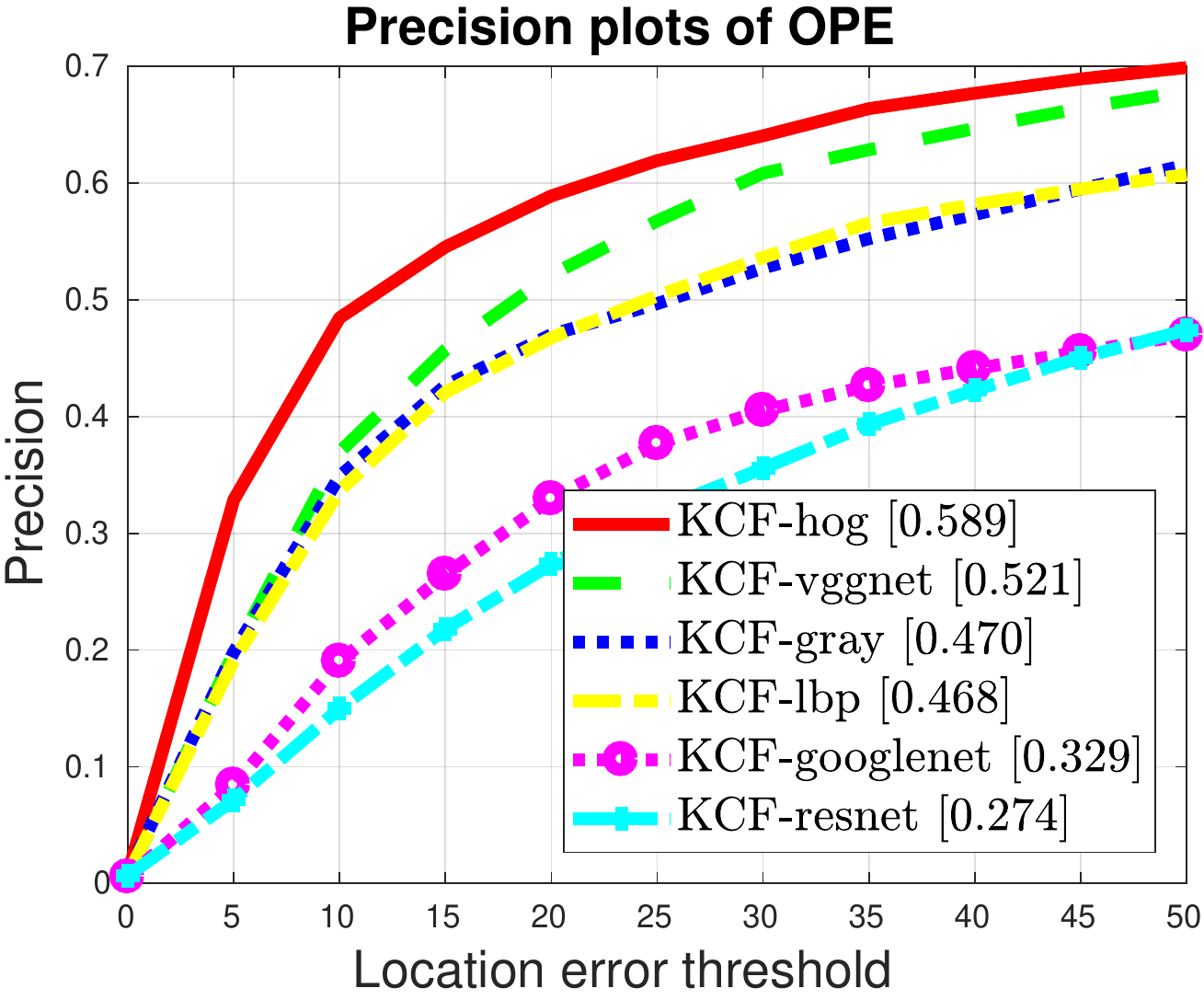}
                \includegraphics[width=0.24\textwidth]{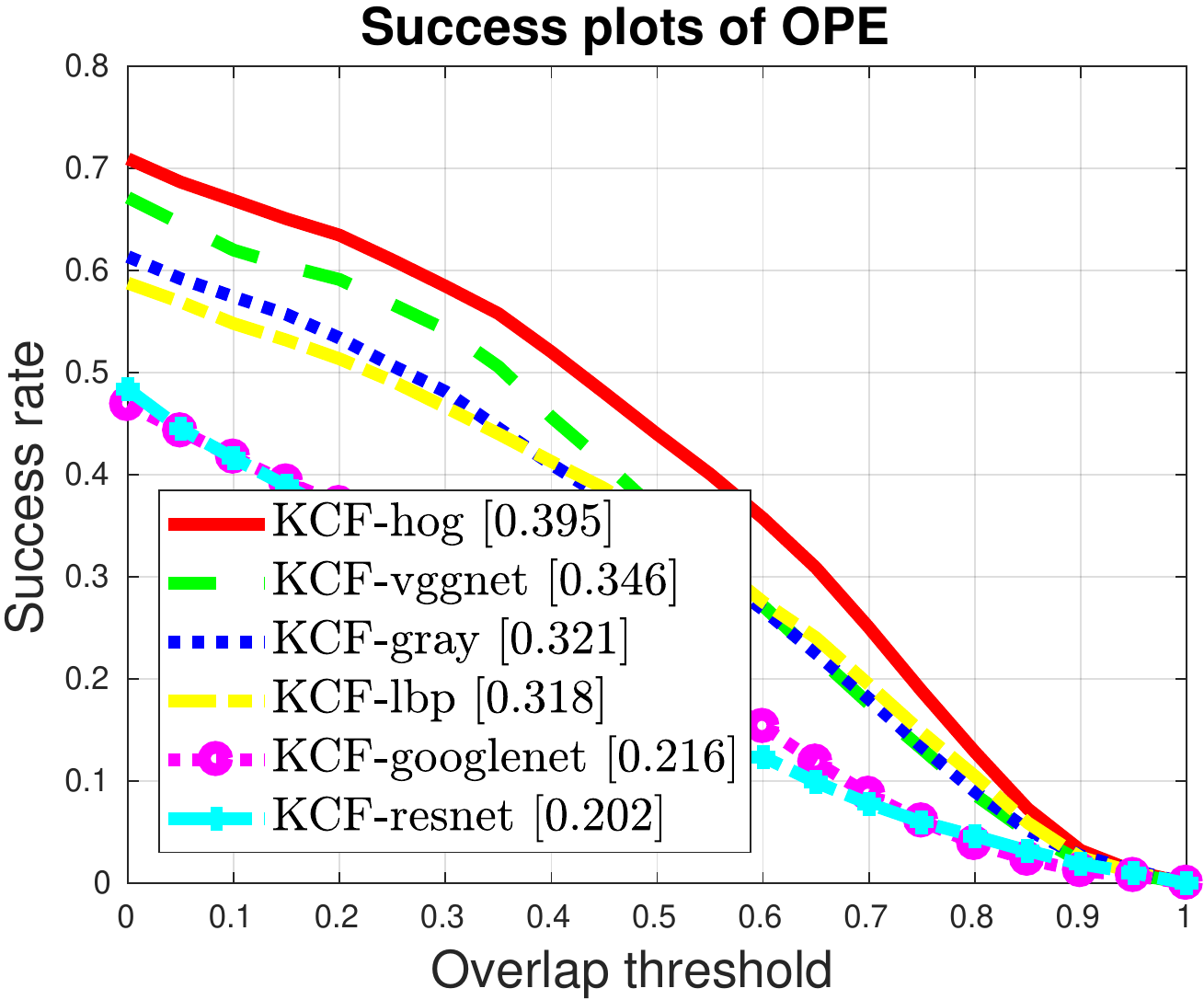}}
  \caption{OPE plots of the {two baseline trackers} using several different feature extractors.}
  \label{featurecomparison}
\end{figure}

To understand the effect of feature extractor on TIR pedestrian tracking performance, we test several different features on two baseline trackers. These features are commonly used in object tracking and detection.
\begin{itemize}
  \item \textbf{Gray.} It just uses pixel values as features.
  \item \textbf{SIFT}~\cite{SIFT}. It is a local feature descriptor and robust to the scale variation. { Here, we use its fast version: dense SIFT using VLFeat toolkit~\cite{vedaldi08vlfeat}}.
  \item \textbf{Haar-like}~\cite{haar}. It reflects the gray variation of the image. We use the simplest form, rectangular Haar-like features.
  \item \textbf{Gabor}~\cite{gabor}. It can capture the texture information of the image using a series of different direction's Gabor filters.
  \item \textbf{LBP}~\cite{LBP}. It is a simple but efficient local texture descriptor, which has two advantages including rotation invariance and gray invariance.
  \item \textbf{HOG}~\cite{HOG}. It is used to capture the local gradient direction and gradient intensity distribution of the image.
  \item {\textbf{Deep feature.} It is extracted from a pre-trained deep network such as VGGNet-19~\cite{VGG-Net}, GoogLeNet~\cite{GoogLeNet}, and ResNet-101~\cite{ResNet}. We use the last convolutional layer feature of them for testing. }
\end{itemize}
We use ridge regression~\cite{wang2015understanding} {and KCF~\cite{KCF} as baseline trackers} and the results are shown in Fig.~\ref{featurecomparison}.

\vspace{3mm}
\noindent{\textbf{Analysis.}} Fig.~\ref{RR} shows that the baseline tracker ridge regression using the HOG feature achieves the best performance higher than the one using the gray feature by about $20\%$. It demonstrates that local gradient features are more helpful for TIR pedestrian tracking. Although the gray features are commonly used in the previous studies, it has a low discriminative ability for the TIR pedestrian object. The assumption that the object is warmer than its background is no longer suitable for TIR pedestrian tracking, because TIR pedestrians often have similar intensity to their backgrounds. In addition, we can see that the local texture feature LBP obtains the second-best performance and its performance is close to the best performance. This illustrates that along with the improvement of the thermal image's quality and resolution, texture features are useful for TIR pedestrian tracking. { Fig.~\ref{KCF} shows that the baseline tracker, KCF, using the HOG feature obtains the better performance than the one using three different deep features. This indicates that these pre-trained networks are not suitable for TIR pedestrian tracking when simply using the single convolutional layer feature. We think that these deep networks, trained on RGB images with classification task, lack the discriminative capacity to the TIR pedestrian object. Furthermore, we can see that tracking performance becomes worse when the deep networks get deeper. We suggest that the deeper convolutional layer feature lacks spatial information, which is not conducive to the precise localization of the object. However, the tracking performance gets much better when multiple convolutional layer features are combined (see MCFTS of Fig.~\ref{Overallperformance}), as illustrates that deep features have the potential capacity to obtain better tracking results when used properly. }

\vspace{3mm}
\noindent{\textbf{Findings.}} {From the comparison of Fig.~\ref{featurecomparison}, Fig.~\ref{motioncomparison} and Fig.~\ref{observationcomparison}, we can see that feature extractor is the most important component in the TIR pedestrian tracker.} It has a major effect on the TIR pedestrian tracker's performance. Choosing or developing a strong feature can dramatically enhance tracking performance.

\subsection{Motion Model.}
\label{MM}
\begin{figure}
  \centering
   \subfloat[{(a) Gray}]{\includegraphics[width=0.24\textwidth]{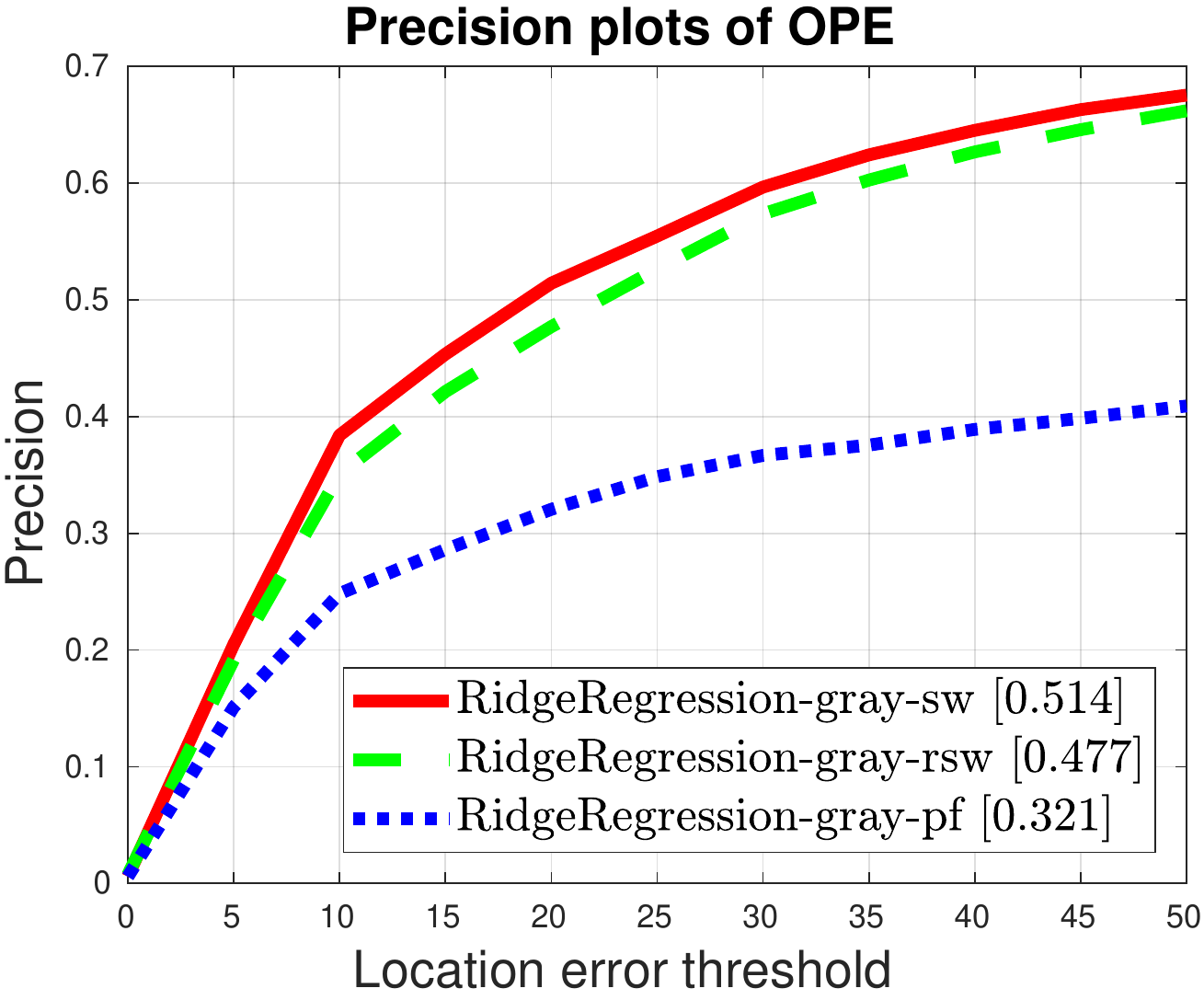}
               \includegraphics[width=0.24\textwidth]{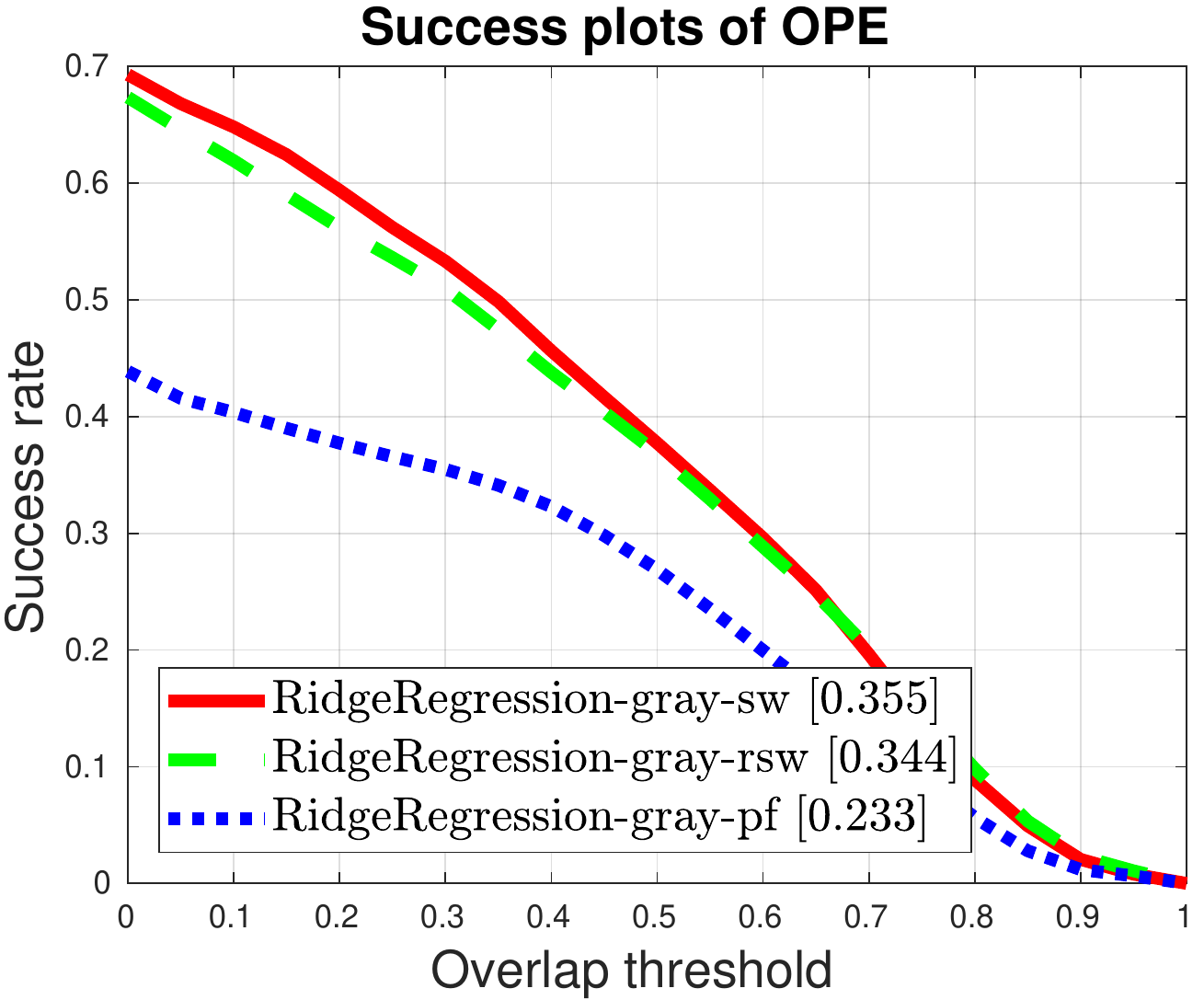}} \vspace{-0.1in} \\
     \subfloat[{(b) HOG}]{\includegraphics[width=0.24\textwidth]{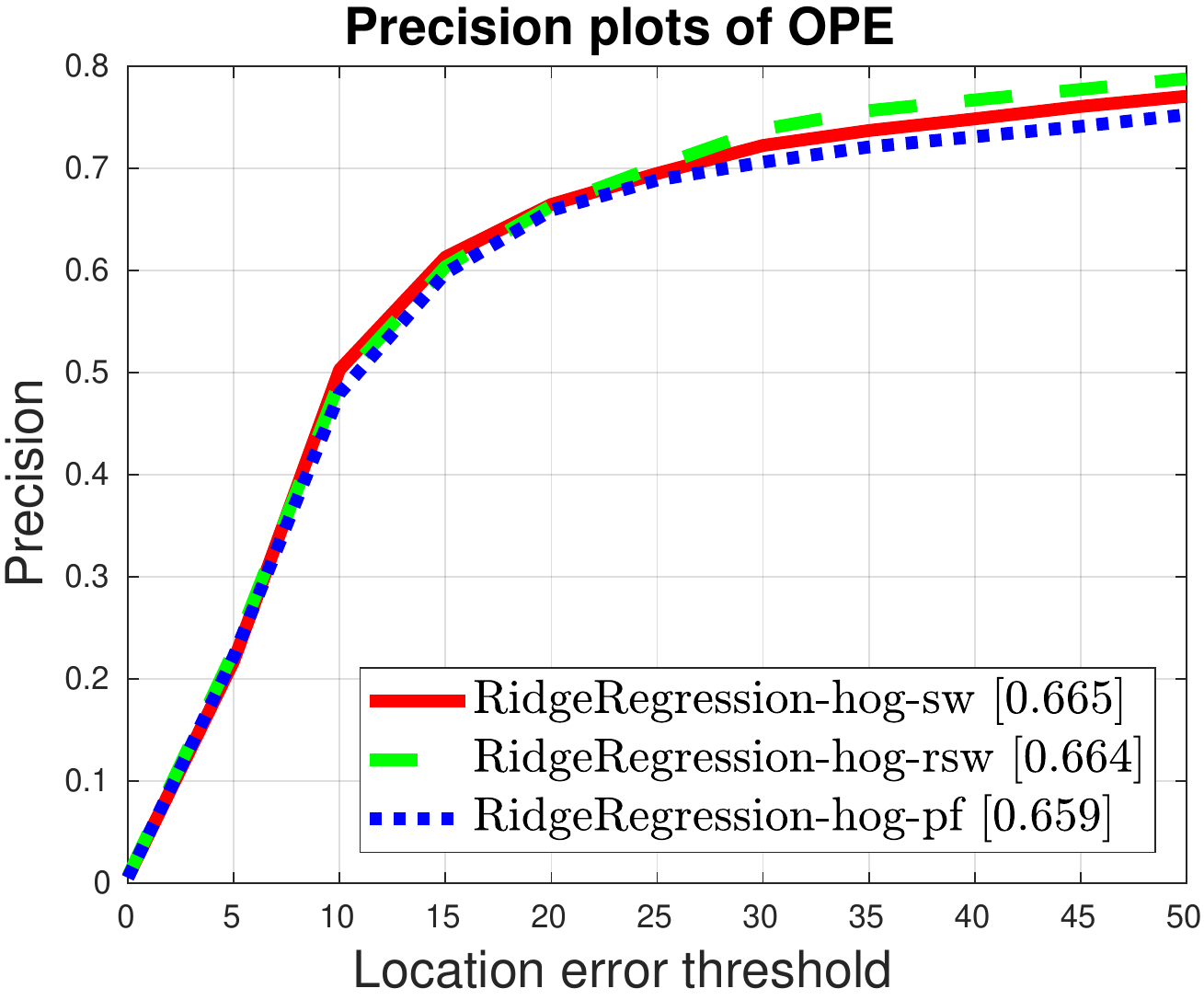}
                 \includegraphics[width=0.24\textwidth]{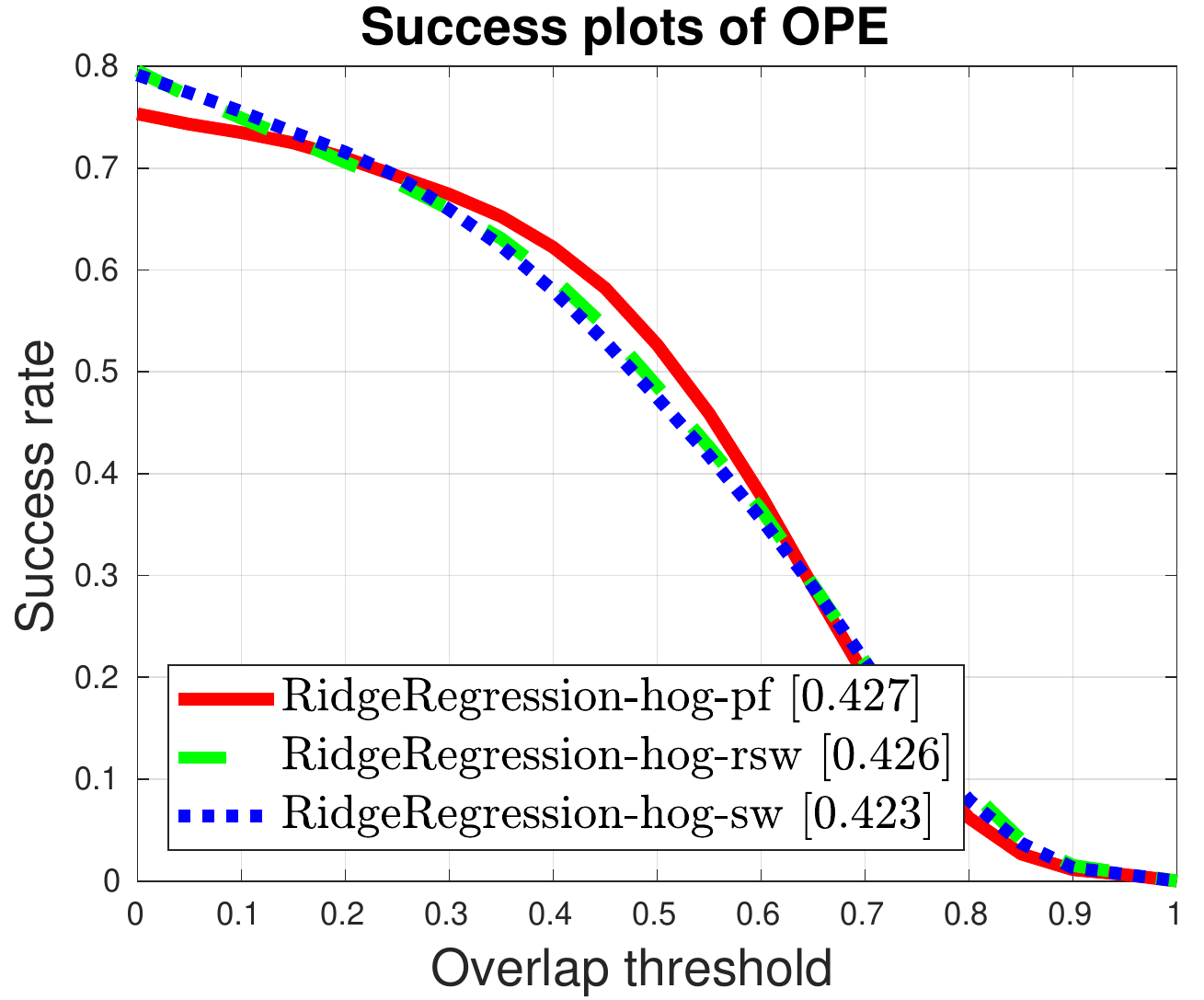}}
  \caption{OPE plots of the baseline tracker using three different motion models with two different feature extractors.
  The abbreviations pf, sw, and rsw denote the particle filter, sliding window, and radius sliding window respectively.}
  \label{motioncomparison}
\end{figure}

To understand the effect of motion model on TIR pedestrian tracking performance, we test three commonly used motion models on a baseline tracker with two different features.
\begin{itemize}
  \item \textbf{Particle Filter}~\cite{particlefilter}. It is a sequential Bayesian importance sampling technique which belongs to the stochastic search method.
  \item \textbf{Sliding Window.} Sliding window is a kind of exhaustive search method. It simply considers all candidates within a square neighborhood.
  \item \textbf{Radius Sliding Window}~\cite{hare2016struck}. Radius sliding window is an improved version of the sliding window. It just considers all candidates within a circular region.
\end{itemize}
We still use ridge regression~\cite{wang2015understanding} as the baseline tracker and the results are shown in Fig.~\ref{motioncomparison}.

\vspace{3mm}
\noindent{\textbf{Analysis.}}  Particle filter has several advantages, e.g., it can solve the scale variation problem and recover from the tracking failure. However, from Fig.~\ref{motioncomparison}, it is easy to see that the particle filter is much worse than the sliding window when we use the weak feature (Gray) in the baseline tracker. We suggest that this is mainly because the gray feature lacks the discriminative capacity leading to the drift of the particle filter.
On the contrary, when we use the strong feature (HOG) in the baseline tracker, three motion models perform the same. It is interesting that the particle filter performs with no apparent superiority on tracking performance even though it involves a high computation complexity.

\vspace{3mm}
\noindent{\textbf{Findings.}} Different motion models have a minor effect for the tracker's performance when the feature is strong enough. Therefore, a faster search strategy (e.g., sliding window) is more helpful for TIR pedestrian tracking.

\subsection{Observation Model.}
\label{OM}
\begin{figure}
  \centering
   \subfloat[{(a) Gray}]{\includegraphics[width=0.24\textwidth]{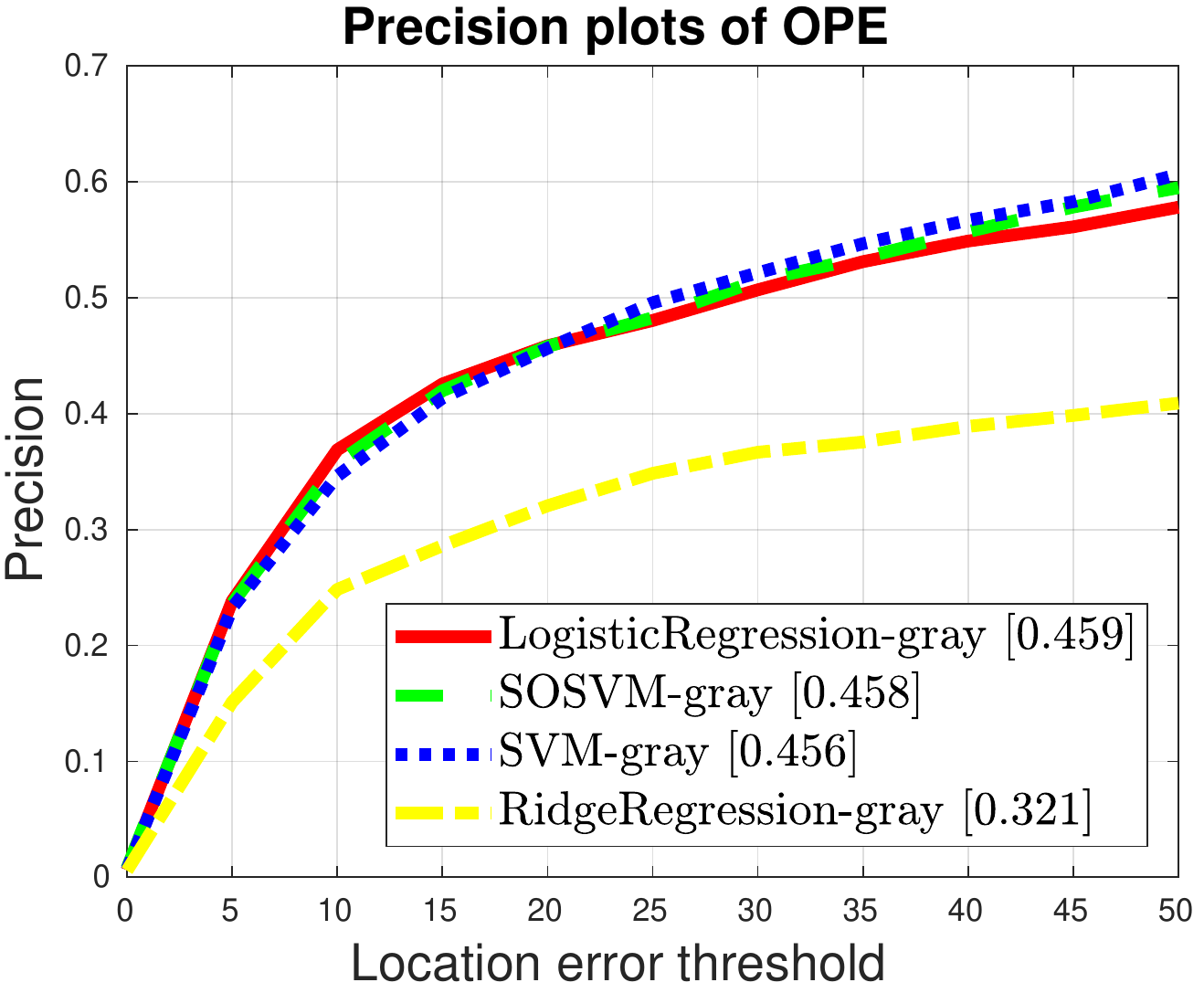}
                \includegraphics[width=0.24\textwidth]{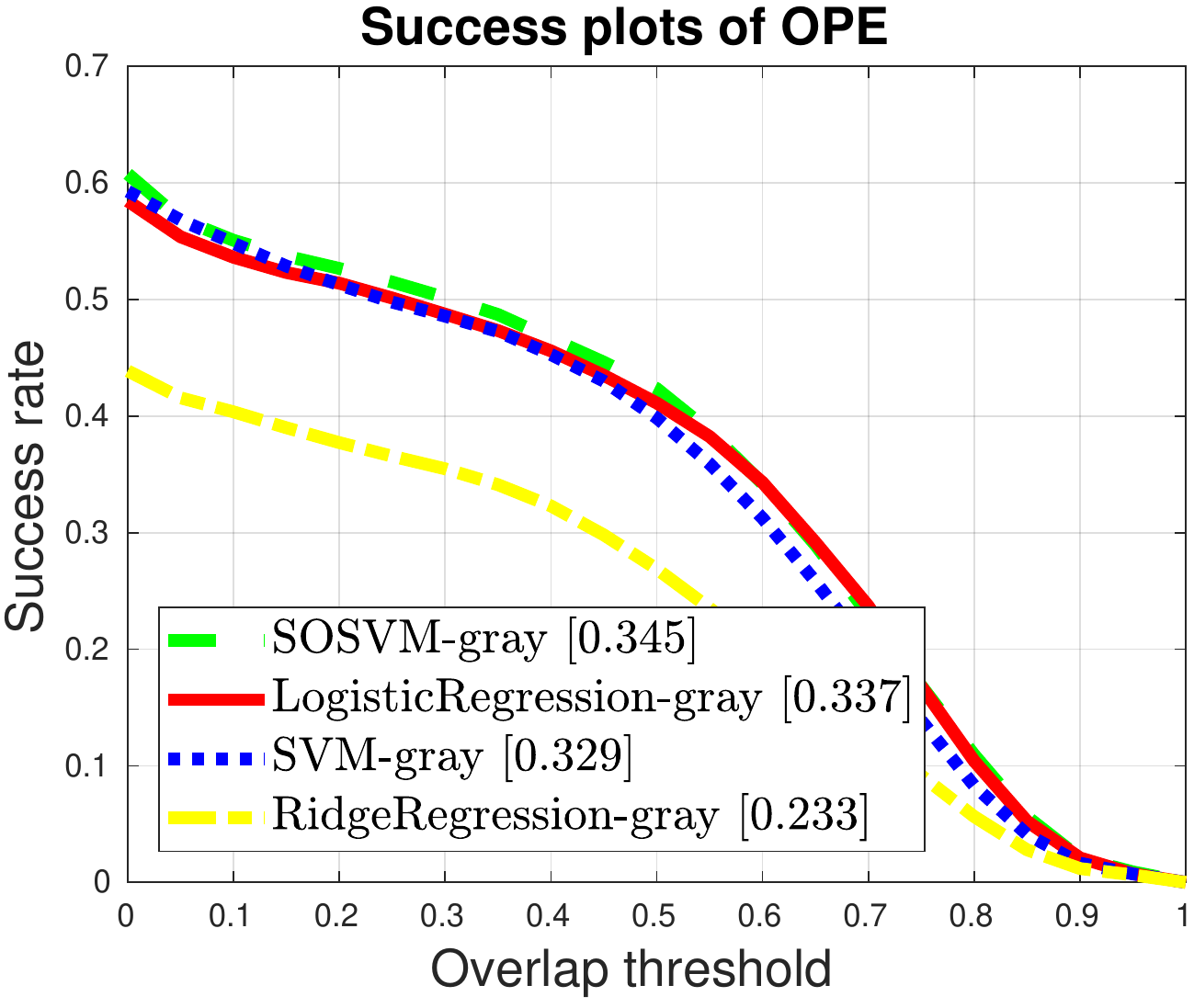}}\vspace{-0.1in}\\
     \subfloat[{(b) HOG}]{\includegraphics[width=0.24\textwidth]{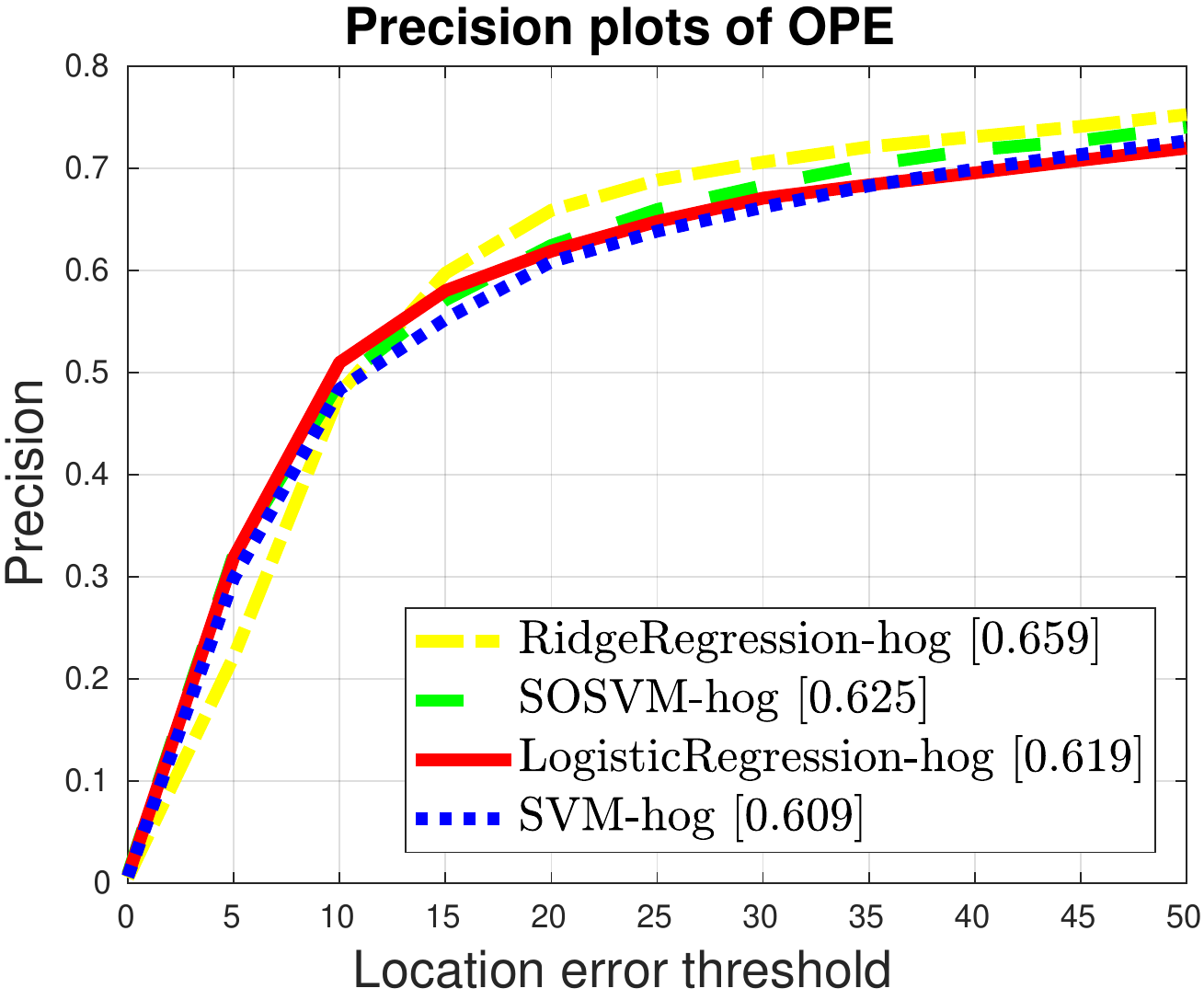}
                 \includegraphics[width=0.24\textwidth]{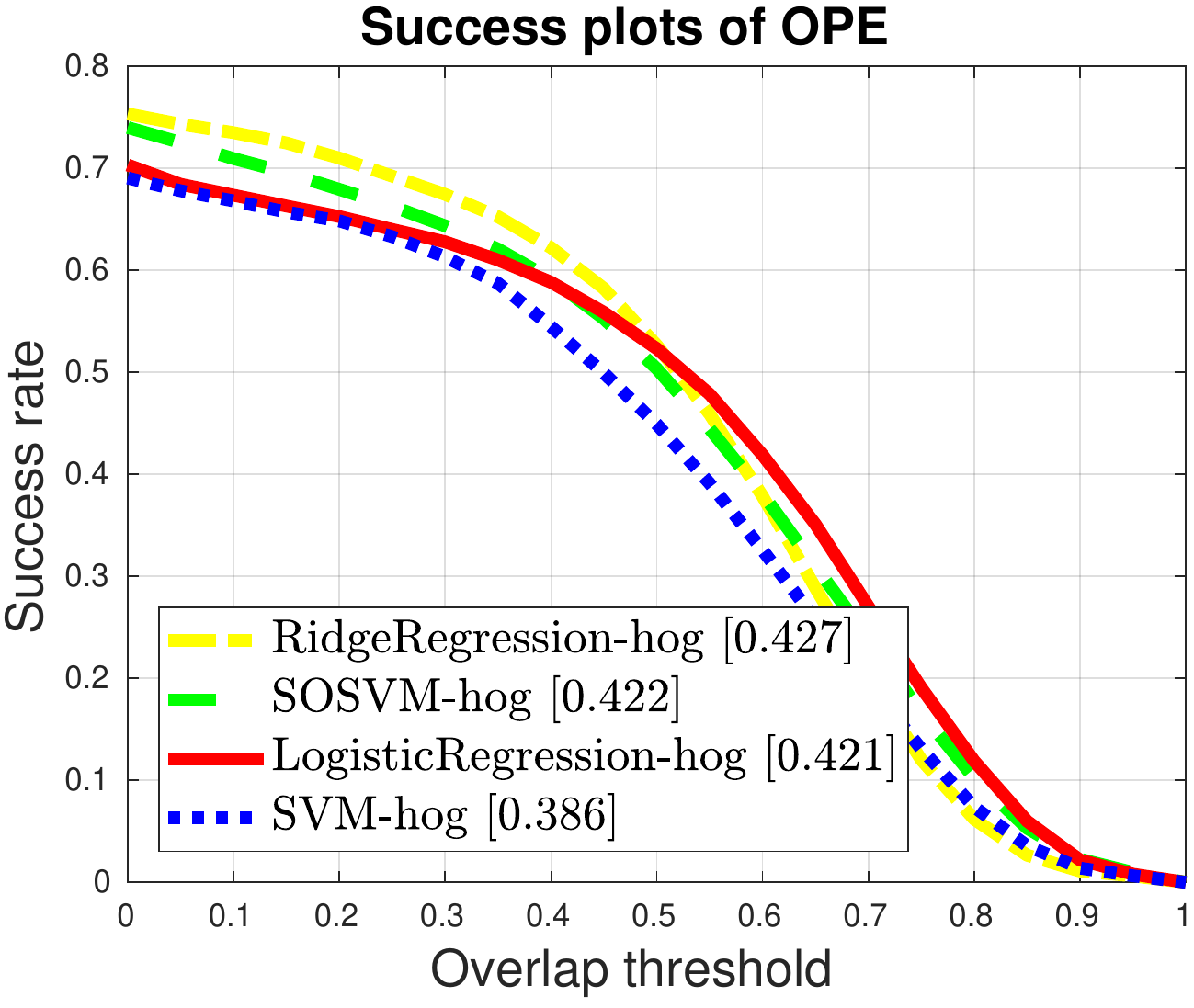}}
  \caption{OPE plots of the four trackers using different observation models with two different feature extractors.}
  \label{observationcomparison}
\end{figure}

Observation models are widely studied in the tracking filed since the rapid  development of the machine learning method. To understand the effect of observation model on TIR pedestrian tracking performance, we test four observation models using two different features.
\begin{itemize}
  \item \textbf{Logistic Regression.} It is a kind of linear regression with $\ell_{2}$ regularization. We use gradient descent to update the model online.
  \item \textbf{Ridge Regression.} Least squares regression with $\ell_{2}$ regularization is used. The method comes from~\cite{wang2015understanding}.
  \item \textbf{SVM.} It is a standard SVM with hinge loss and $\ell_{2}$ regularization.
  \item \textbf{Structured Output SVM (SOSVM).} This method is an enhanced version of SVM and comes from~\cite{hare2016struck}.
\end{itemize}
These four observation models are ranked according to their classification ability. The results are shown in Fig.~\ref{observationcomparison}.

\vspace{3mm}
\noindent{\textbf{Analysis.}} Fig.~\ref{observationcomparison} shows that a strong observation model SOSVM achieves the best performance when the weak feature (Gray) is used. It exceeds the weak observation model, ridge regression, by more than $10\%$. However, when we use the strong feature HOG, the weak observation model ridge regression obtains the best performance against the stronger SVM and SOSVM. Furthermore, it is easy to see that these observation models have similar performance when the strong feature is used. Similar observations are reported in visual tracking~\cite{wang2015understanding}.

\vspace{3mm}
\noindent{\textbf{Findings.}} The strong observation model can obtain higher tracking performance when the feature is weak. However, when the feature is strong enough, different observation models have a minor gap in tracking performance.

\section{Conclusion and Future Work}
\label{conclusion}
In this paper, we develop a TIR pedestrian tracking benchmark dataset for TIR pedestrian tracker evaluation. A large scale evaluation experiment is carried out on our benchmark with nine publicly available trackers. Based on our evaluation  results and analysis, several observations are highlighted for understanding the TIR pedestrian tracker. First, the scale estimation strategy is very important for TIR pedestrian tracking, and it can greatly improve tracking performance. Second, the background information is crucial for a discriminative tracker, and it can enhance the discriminative power of the model. Third, deep learning based trackers have the potential to obtain superior performance in the TIR pedestrian tracking.

In addition, in order to better understand the TIR pedestrian tracking, we conducted three validation experiments on each component of the tracker. Some interesting findings are helpful for future research. First, the feature extractor is the most important component in the TIR pedestrian tracker; a strong feature can significantly improve tracking performance. Second, different motion models have a minor effect on the tracking results when the feature is strong enough. Third, different observation models also have a minor effect on tracking performance when the feature is strong.  On the contrary, when the feature is weak, a stronger observation model often can achieve better performance.

The evaluation and validation { of experimental results help us understand}  several aspects of the TIR pedestrian tracker. This will promote the development of this field. In the future, we are going to extend the dataset to include more thermal sequences and explore more challenge factors in TIR pedestrian tracking.

\section*{Acknowledgment}
This research was supported by the National Natural Science Foundation of China (Grant No.61672183, 61502119), by the Shenzhen Research Council (Grant Nos. JCYJ20170815113552036,
JCYJ20170413104556946, JCYJ20160406161948211, JCY-J20160226201453085), and by the Natural Science Foundation of Guangdong Province (Grant No. 2015A030313544).




%
\bibliographystyle{IEEEtran}


%
%

%
%
\begin{IEEEbiography}[{\includegraphics[width=1in,height=1.25in,clip,keepaspectratio]{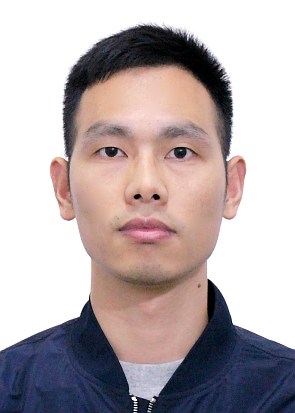}}]{Qiao Liu}
received the B.E degree in computer science from the Guizhou Normal University, Guiyang, China, in 2016. He is pursuing the Ph.D. degree with the Department of Computer Science and Technology, Harbin Institute of Technology Shenzhen Graduate School, China. His current research interests include thermal infrared object tracking and machine learning.
\end{IEEEbiography}


\begin{IEEEbiography}[{\includegraphics[width=1in,height=1.25in,clip,keepaspectratio]{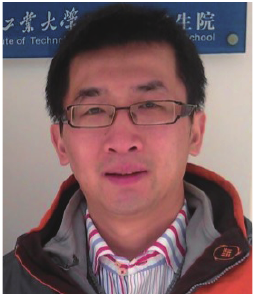}}]{Zhenyu He}
received his Ph.D. degree from the Department of Computer Science, Hong Kong Baptist University, Hong Kong, in 2007. He is currently a Professor with the School of Computer Science and Technology, Harbin Institute of Technology Shenzhen Graduate School, China. His research interests include sparse representation and its applications, deep learning and its applications, pattern recognition, image processing, and computer vision.
\end{IEEEbiography}

\begin{IEEEbiography}[{\includegraphics[width=1in,height=1.25in,clip,keepaspectratio]{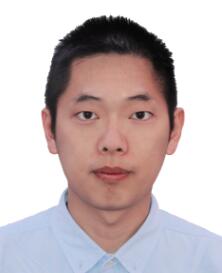}}]{Xin Li}
received the B.E degree in 2014 and currently, he is a Ph.D candidate with the Department of Computer Science both from Harbin Institute of Technology Shenzhen Graduate School. His research interests lie primarily in visual tracking, image processing and machine learning.
\end{IEEEbiography}
\vfill

\newpage
\begin{IEEEbiography}[{\includegraphics[width=1in,height=1.25in,clip,keepaspectratio]{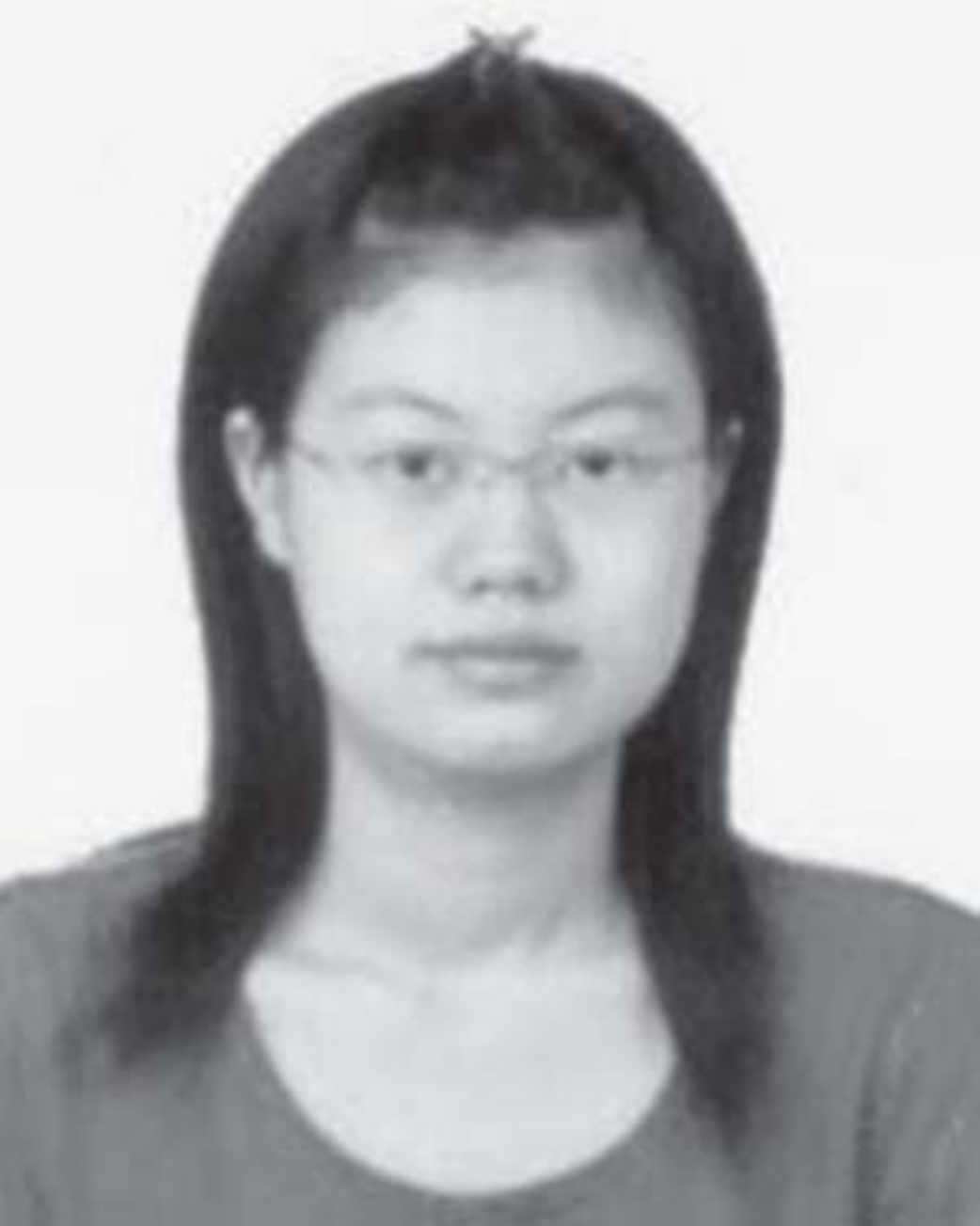}}]{Yuan Zheng}
is currently an Assistant Professor at the College of Computer Science, Inner Mongolia University, Hohhot, China. Before that, she was a postdoctoral researcher at the School of Computer Science, Harbin Institute of Technology Shenzhen Graduate School, China. She received a Ph.D. degree in the Institute of Automation, Chinese Academy of Sciences, Beijing, China, in 2014.  She received the B.S. and M.S. degrees in the Harbin Institute of Technology (HIT), Harbin, China, in 2006 and 2008, respectively. Her main research interests include video surveillance, object analysis, and camera calibration.
\end{IEEEbiography}

\vfill


\end{document}